\documentclass[times, review, 10pt]{elsarticle}
\usepackage[numbers]{natbib}
\usepackage{float}
\usepackage{graphicx}
\usepackage{subfigure}
\usepackage{diagbox} 
\usepackage{soul}

\usepackage{xcolor}

\definecolor{teal}{rgb}{0.0, 0.5, 0.5}

 \usepackage{amsmath,amssymb,amsfonts}

 \usepackage{hyperref}

\hypersetup{
    colorlinks=true,
    linkcolor=blue,
    filecolor=blue,      
    urlcolor=blue,
    citecolor=blue,
}

\usepackage{caption}
\def\,{$\mskip\thinmuskip$} \def\!{$\mskip-\thinmuskip$}

\usepackage{algorithm}
\usepackage[algo2e]{algorithm2e} 
\usepackage{algorithmicx}

\def\BibTeX{{\rm B\kern-.05em{\sc i\kern-.025em b}\kern-.08em
    T\kern-.1667em\lower.7ex\hbox{E}\kern-.125emX}}

\begin{document}
\title{Preserving logical and functional dependencies in synthetic tabular data}


\author[1]{Chaithra Umesh}
\author[1]{Kristian Schultz}
\author[1]{Manjunath Mahendra}
\author[1,4]{Saptarshi Bej}
\author[1,2,3]{Olaf Wolkenhauer}

\affiliation[1]{Institute of Computer Science, University of Rostock; Germany}
\affiliation[2]{Leibniz-Institute for Food Systems Biology; Technical University of Munich, Freising; Germany}
\affiliation[3]{Stellenbosch Institute for Advanced Study, South Africa}
\affiliation[4]{Indian Institute of Science Education and Research, Thiruvananthapuram, India}


\begin{abstract}
Dependencies among attributes are a common aspect of tabular data. However, whether existing tabular data generation algorithms preserve these dependencies while generating synthetic data is yet to be explored. In addition to the existing notion of functional dependencies, we introduce the notion of logical dependencies among the attributes in this article. Moreover, we provide a measure to quantify logical dependencies among attributes in tabular data. Utilizing this measure, we compare several state-of-the-art synthetic data generation algorithms and test their capability to preserve logical and functional dependencies on several publicly available datasets. We demonstrate that currently available synthetic tabular data generation algorithms do not fully preserve functional dependencies when they generate synthetic datasets. In addition, we also showed that some tabular synthetic data generation models can preserve inter-attribute logical dependencies. Our review and comparison of the state-of-the-art reveal research needs and opportunities to develop task-specific synthetic tabular data generation models.
\end{abstract}

\maketitle


\section{Introduction}\label{sec:introduction}

\textbf{Dependencies among attributes are a common aspect of tabular data.} For instance, two attributes, such as Gender and Pregnancy, in some clinical data have a clear dependency in a logical sense because it is not possible for a male to be pregnant. A well-known fact in Database Management Systems is that if one wants to remove redundancies by dividing larger tables into smaller ones (Normalization) \cite{liu_discover_2012}, one needs tools to identify functional dependencies present among the attributes of the larger table \cite{yao_mining_2008}. Preserving functional dependencies in synthetic tabular data is an area that has not been explored. Dependencies exist in both tabular and image data. A recent study by Tongzhen Si \textit{et al.}\@, published in Pattern Recognition, discusses capturing inter-image dependencies between pedestrian images and intra-pixel dependencies within each image using attention mechanism \cite{si_spatial-driven_2022}.\par

\textbf{Functional dependencies (FD).} Functional dependency describes a relationship between attributes (columns/features) in a table. For two functionally dependent features, for each value of one feature, there is a unique value for the other feature \cite{wei_towards_2023}. Given a tabular dataset $T$ with values $T_{i, j}$ for row $i$ in $R = \{ 1, 2, 3, \ldots, n \}$ and column $j$ in $C = \{ 1, 2, 3, \ldots, m \}$.
Let $\mathcal{A} = (\alpha_1, \alpha_2, \alpha_3, \ldots )$ be a subset of columns with $\alpha_k \in C$.
The term $T_{i, \mathcal{A}} = (T_{i, \alpha_1}, T_{i, \alpha_2}, T_{i, \alpha_3}, \ldots)$ denotes the tuple that represents the values of row $i$ for the selected columns in $\mathcal{A}$.
The set $A = \{ T_{i, \mathcal{A}} \colon i \in R \}$ is the set of all tuples that exists in the table for the given selection of columns $\mathcal{A}$.
In the same way, we have a second selection of columns $\mathcal{B}$ and the corresponding set of tuples $B = \{ T_{i, \mathcal{B}} \colon i \in R \}$.

Let be $a \in A$ and $b \in B$.
Our table $T$ implies a relation $\sim_T \subseteq A \times B$ where $(a, b)$ is in $\sim_T$ if and only if there exists an $i \in R$ that $T_{i, \mathcal{A}} = a$ and $T_{i, \mathcal{B}} = b$ \cite{khamis_learning_2020}.
For short we write $a \sim_T b$ when $(a, b)$ is in $\sim_T$.

A functional dependency between $A$ and $B$ is denoted as $A \to B$ ($B$ is functionally dependent on $A$), meaning that for every value $a$ in $A$, there is a unique value $b$ for that $a \sim_T b$ holds \cite{bernstein_synthesizing_1976}. We can write:
\[
   (A \to B) \Leftrightarrow \big( \forall i_1, i_2 \in R : T_{i_1, \mathcal{A}} = T_{i_2, \mathcal{A}} \implies T_{i_1, \mathcal{B}} = T_{i_2, \mathcal{B}} \big)
\]
        
In Table \ref{FD example}, an example of functional dependency is that the disease uniquely determines the examiner. Another example is the association between DNA segments and diseases \cite{liu_discover_2012}. Identifying functional dependencies plays a role in diagnostics, prognostics, and therapeutic decisions.\par

The literature provides several algorithms for extracting functional dependencies for a given data \cite{sug_comparison_2022, caruccio_discovering_2021, yao_mining_2008}. However, research in the field of synthetic data has yet to explore how well synthetic data generation algorithms can preserve functional dependencies in synthetic data compared to real data.

\begin{table*}[!ht]
    \footnotesize
    \begin{center}
     \begin{tabular}{c|c|c|c} 
      \hline
      
       {\textcolor{red}{Disease}}&  {\textcolor{red}{Examiner}} &  {\textcolor{teal}{Pregnant}} &  {\textcolor{teal}{Gender}}\\
       \hline
       
       {Heart failure} & {Cardiologist} & {No} & {M}  \\
       \hline
      
       {Tuberculosis} & {Pulmonologist} & {Yes}  & {F}  \\
       \hline
    
       {Heart failure} & {Cardiologist} & {No}  & {M}  \\
       \hline
    
       {Heart failure} & {Cardiologist} & {No}  & {F}  \\
       \hline 
    
    \end{tabular}
    \caption{An example of clinical data with functional and logical dependencies. The first and second columns highlighted in red display functional dependency across all instances, while the third and fourth columns in green denote logical dependency, particularly when Gender is M and Pregnancy status is No. For two functionally dependent features, for each value of one feature, there is a unique value for the other feature.}\label{FD example}
    \end{center}
\end{table*}

\textbf{Logical dependencies} describe how one attribute's value can logically determine another attribute's value.
Let's collect all values of $B$ that are in the same row as a given $a \in A$ with $D(a) = \{ b \in B \colon a \sim_T b \}$.
So $D(a)$ is the set of all $b$ in $B$ that are dependent of a given $a$.
Obviously, we have that $D(a)$ is a subset of $B$ or $D(a)$ is equal to $B$.
Assume there are some $b \in B$ that are missing in $D(a)$.
Then we can tell by knowing the value for $a = T_{i, \mathcal{A}}$ for a chosen row $i$ that these $b \in B \setminus D(a)$ will never occur in this particular $T_{i, \mathcal{B}}$.
The smaller the set $D(a)$, the better our prediction for possible $b$ will be.
In the case of $|D(a)| = 1$ for all $a \in A$, we have a functional dependency of $A \to B$ given by our table.

Table \ref{FD example} illustrates a logical dependency between the Gender being M(Male) and Pregnancy being No.
When certain conditions consistently hold across different categories, a logical dependency exists between them.
It is important to note that if attributes are functionally dependent, they are also naturally logically dependent. \par

When we create synthetic data, it is important to respect logical dependencies, for example, to avoid generating synthetic tables with males being pregnant. In contrast to the well-explored concept of functional dependencies, the idea of logical dependencies remains largely unexplored for synthetic data.

\textbf{Measures to quantify logical dependencies among attributes.} Currently, no standard measures are available to determine the logical dependencies between attributes in a given dataset. In Section \ref{Q_function}, we introduce a novel measure to identify inter-attribute logical and functional dependencies.

\begin{figure}[ht]%
    \centering

    {{\includegraphics[width=3.5cm]{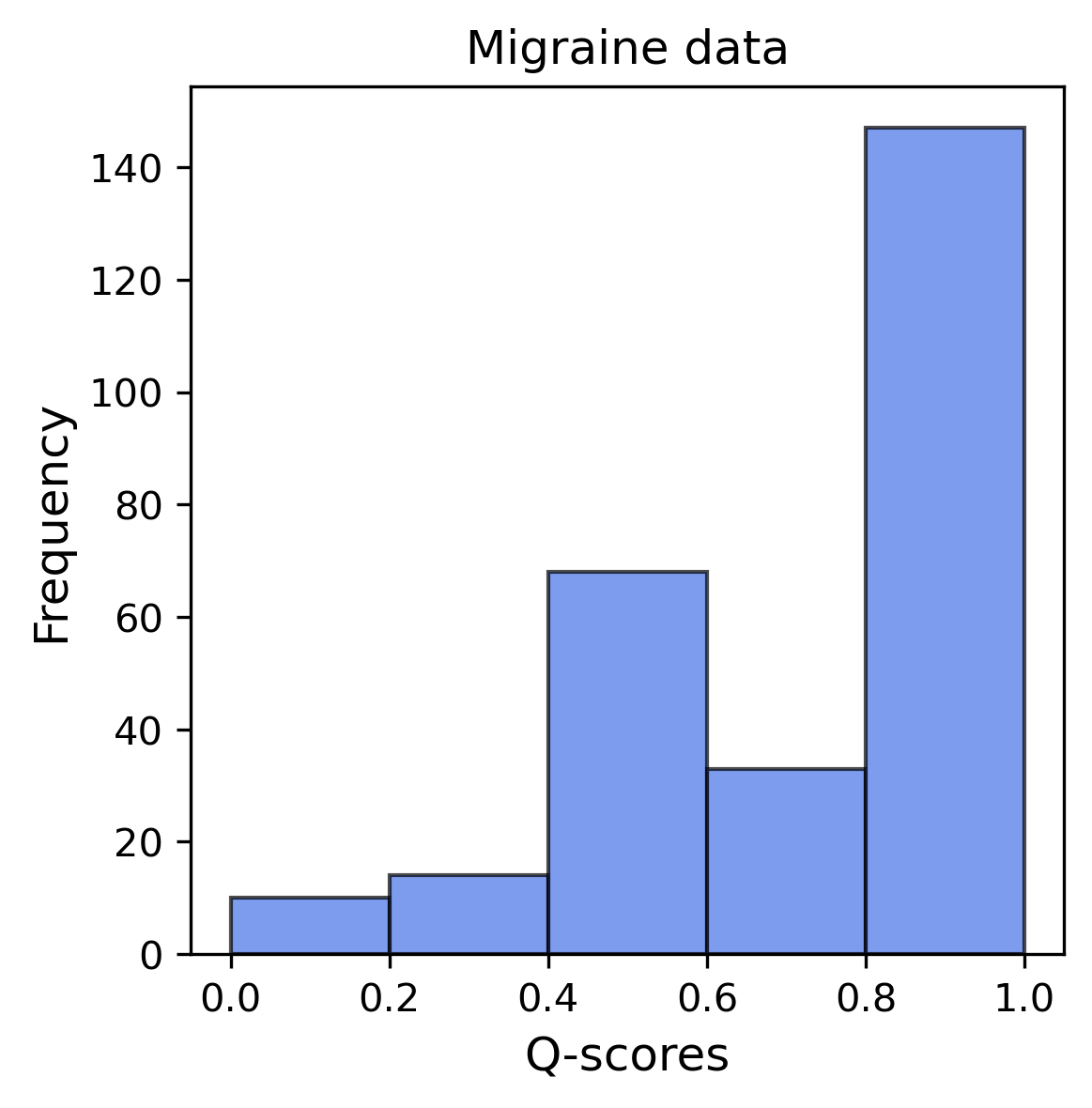} }\label{m}}%
    \qquad
    {{\includegraphics[width=3.5cm]{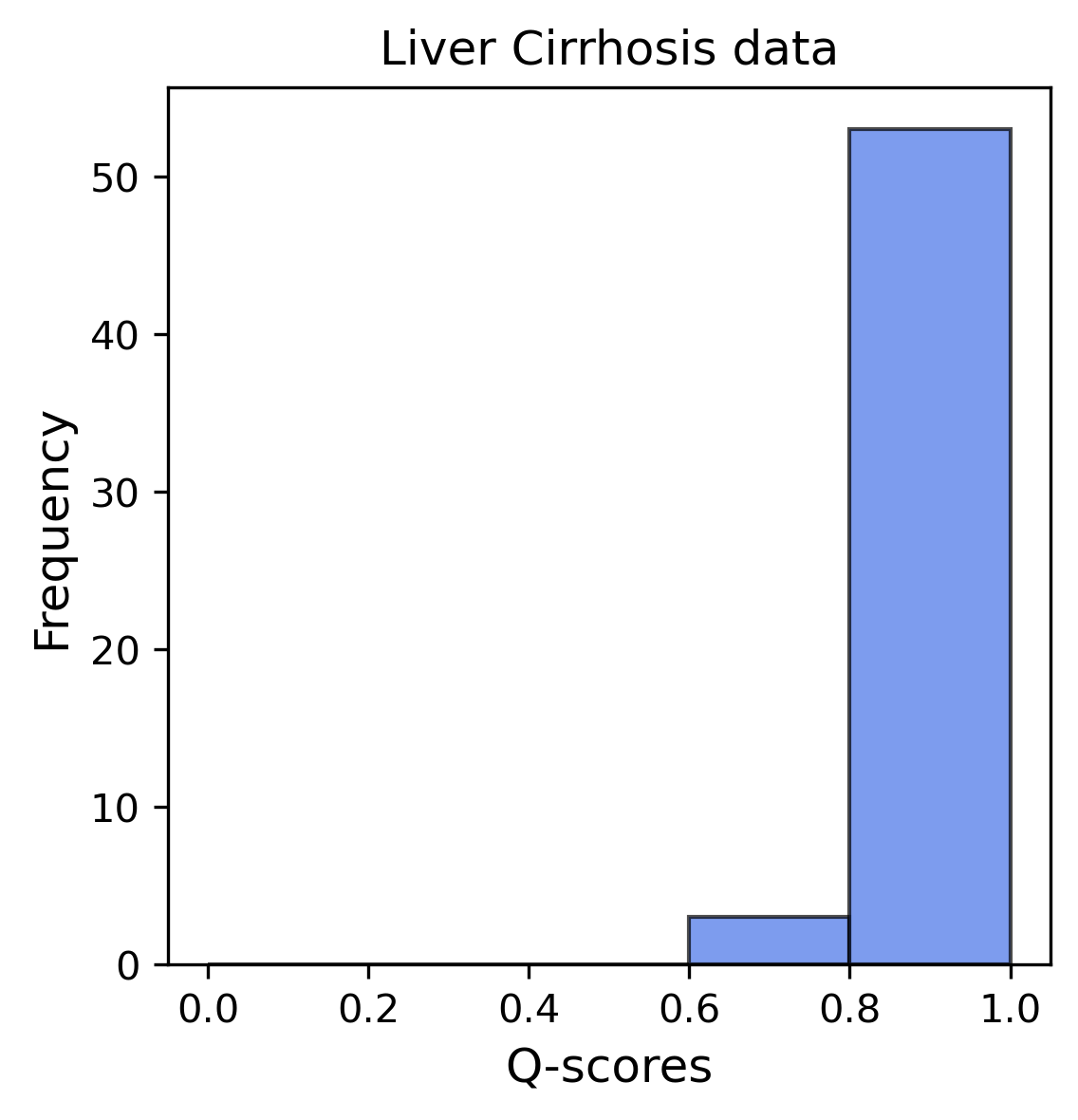} }\label{a}}%
    \qquad
    {{\includegraphics[width=3.5cm]{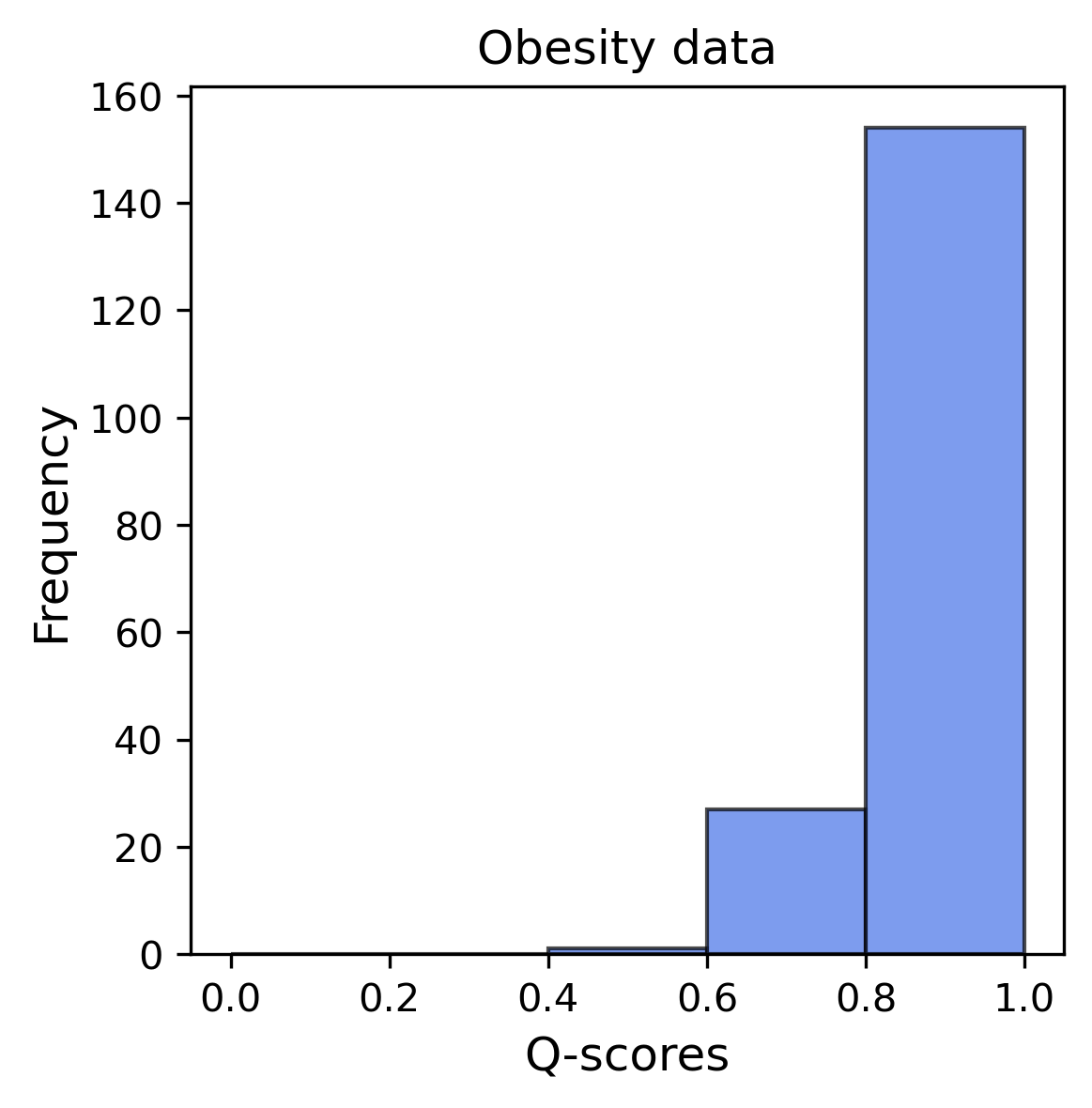}}\label{o}}%
    \qquad
    {{\includegraphics[width=3.5cm]{Data_distributions_obesity.png}}\label{s}}%
    \qquad
    {{\includegraphics[width=3.5cm]{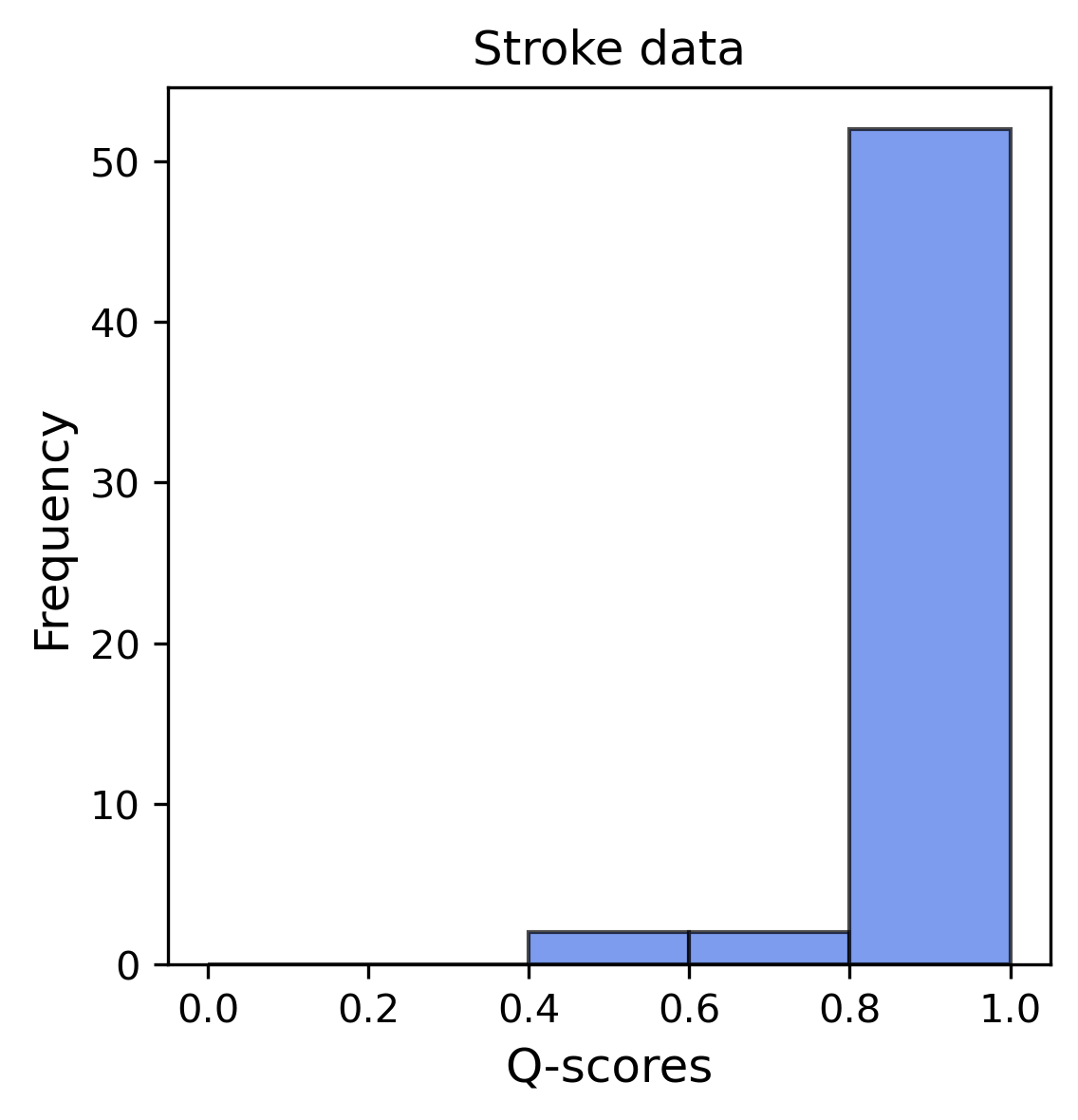}}\label{l}}%
    \caption{\small The figure above has five histograms corresponding to the five datasets used in this study. On the $x$-axis, we plot the $Q$-scores in discrete bins. On the $y$-axis, we record the number of attribute pairs that attain a certain Value. For a pair of attributes, a $Q$-score value between $0$ and $1$ implies logical dependency between those two attributes, while a $Q$-score value of exactly $0$ indicates functional dependency between those two attributes. Conversely, a $Q$-score value of $1$ signifies no dependency between a pair of attributes/features in the data. The Migraine and Airbnb datasets have logical and functional dependencies, while other datasets only have logical dependencies between attributes in real data.}
    \label{q-score-distribution}
\end{figure}

Defining with $C^* = \{ (\alpha_1, \ldots ) \colon \{ \alpha_1, \ldots \} \subseteq C \}$ the set that contains all possible selection of columns. The $Q$-function $Q_T : C^* \times C^* \to [0,1]$ provides $Q$-scores between $0$ and $1$ for every pair of column selections $\mathcal{A} \in C^*$ and $\mathcal{B} \in C^*$ within the dataset. In our study, we concentrate on column selections that contain only one column. If there are $k$ attributes, then $(k^{2} - k)$ $Q$-scores are generated. A score of $0$ in the $Q$ function indicates that the attributes are functionally dependent on each other. If the score is $1$, then the attributes are not dependent on each other. The attributes are logically dependent if the score is between $0$ and $1$. We refer to Section \ref{Q_function} for a detailed mathematical explanation of the $Q$-function. Figure \ref{q-score-distribution} presents the results of the $Q$-function for all the datasets used in the study. 

\begin{figure}[ht]
    \centering
    \includegraphics[width=\textwidth]{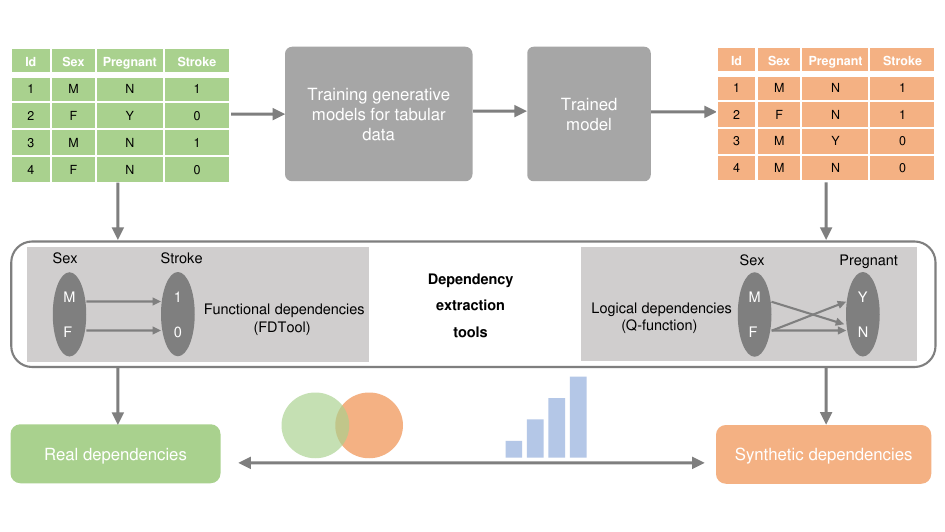}
    \vspace{-0.7cm}
    \caption{Workflow of comparative analysis to assess preservation of functional and logical dependencies in synthetic tabular data using FDTool and Q-function algorithms.}
    \label{workflow}
\end{figure}

Chen \textit{et al.}\@ \cite{chen_faketables_2019} investigated the preservation of functional dependencies in synthetic data generated by GAN-based models and introduced a novel approach emphasizing functional dependencies. However, this approach was limited to bounded real data and did not encompass the entire scope of tabular data \cite{chen_faketables_2019}. Given the importance of logical and functional dependencies in ensuring the authenticity of synthetic data, there is a gap in our understanding of how to preserve these dependencies effectively. Therefore, as a first step in this research direction, an empirical data-driven analysis of the capabilities of synthetic data generation algorithms to retain logical and functional dependencies is essential to enhance the reliability and applicability of synthetic datasets across various analytical domains.\par


In recent years, there has been a significant increase in research efforts focused on generating synthetic tabular data \cite{figueira_survey_2022, xu_modeling_2019, zhao_ctab-gan_2021, kotelnikov_tabddpm_2023}. Along with these models, various performance measures are aimed at evaluating synthetic data's utility, fidelity, and privacy compared to real data \cite{figueira_survey_2022}. Despite the challenges in creating synthetic tabular data, many models have demonstrated their ability to generate data that maintains privacy while preserving its utility \cite{wang_generating_2021, shi_generating_2022}. However, it is worth noting that preserving dependencies among attributes in synthetic data compared to real data can enhance the semantic correctness of the synthetic data.\par
      
In this article, we perform a comparative analysis of synthetic data generation strategies in the context of preservation of logical and functional dependencies on five publicly available datasets, comprising four clinical datasets and one business dataset. We employed seven generative models to create synthetic data for all five datasets. Subsequently, we employed the $Q$-function to extract logical dependencies and FDTool \cite{buranosky_fdtool_2019} to extract functional dependencies from real and synthetic data. Then, we compared the percentage of preserved logical and functional dependencies across all models. There are various publicly available algorithms to extract functional dependencies, and we chose FDTool due to its validation on clinical datasets \cite{buranosky_fdtool_2019}. Our findings reveal that while some generative models preserve logical dependencies to a reasonable extent, none adequately maintain functional dependencies. This study underscores the need for further research to address attribute-dependency preserving generative modeling for tabular data.\par

\section{Related research} 

 We focus on generating synthetic tabular data rather than synthetic data in general. Recently, researchers developed several advanced methods to create realistic tabular datasets \cite{figueira_survey_2022}. Models such as Generative Adversarial Networks (GANs) \cite{goodfellow_generative_2014}, Variational Autoencoders (VAEs) \cite{sami_comparative_2019}, diffusion models \cite{sattarov_findiff_2023}, convex space generators \cite{mahendra_convex_2024} and Large Language Models (LLMs) \cite{zhao_survey_2023} have shown the ability to replicate complex patterns found in real tabular data accurately. Each of these models offers unique approaches and advantages for addressing the specific challenges of tabular data, including handling various data types, preserving relationships between columns, and managing high-dimensional spaces \cite{figueira_survey_2022}. In the following sections, we will explore some of the advanced models we utilized in our experiments.

\subsection{State-of-the-art models employed in the comparative study to generate synthetic tabular data}

    \textit{CTGAN}, introduced by Xu \textit{et al.}\@ in 2019 for synthetic tabular data generation using Conditional Generative Adversarial Networks. The training process incorporates mode-specific normalization to model Non-Gaussian and multimodal continuous distributions \cite{xu_modeling_2019}. A variational Gaussian mixture model (VGM) is employed to determine the number of modes for a continuous column, followed by calculating the probability of each value in the column belonging to a specific mode. This results in each continuous value represented by a one-hot encoded mode and a continuous mode-specific normalized scalar. Discrete data is represented using one-hot encoding \cite{schultz_convgen_2022}. A training-by-sampling strategy addresses the imbalance in categorical features during training to ensure an even sampling of all categories from discrete attributes \cite{xu_modeling_2019}. A conditional vector is employed to select a specific value for each discrete column, with the size of the vector being the sum of the cardinality of each discrete column. The PMF (Probability Mass Function) across all possible values is calculated using the logarithm of the frequency of each value in the column, and the conditional vector is set accordingly once a value is selected in the column \cite{schultz_convgen_2022}. The generator is then provided with this conditional vector and a noise vector containing random values. The idea of ‘packing’ was introduced to prevent mode collapse, where the discriminator works on multiple samples simultaneously \cite{lin_pacgan_2018}. \textit{CTGAN} has been benchmarked with multiple datasets, and the results indicate that it can learn more accurate distributions than Bayesian networks \cite{xu_modeling_2019}.\par

    Zhao \textit{et al.}\@ \cite{zhao_ctab-gan_2021} developed \textit{CTABGAN} to generate synthetic tabular data using Conditional Adversarial Networks. This approach addresses the limitations of previous generative models and incorporates an auxiliary classifier, along with a generator and discriminator, to enhance the integrity of the synthetic data \cite{zhao_ctab-gan_2021}. Unlike \textit{CTGAN}, \textit{CTABGAN} can handle mixed data types, skewed multimode continuous features, and long tail distributions. \textit{CTABGAN} uses a mixed-type Encoder to address mixed data types, treating mixed variables as value pairs consisting of categorical and continuous parts \cite{zhao_ctab-gan_2021}. Additionally, \textit{CTABGAN} includes a training method that ensures each column has an equal probability of being selected and uses a specific mode sampled from the probability distribution (logarithm of the frequency) for continuous variables \cite{zhao_ctab-gan_2021}. Zhao \textit{et al.}\@ pre-process variables with a logarithm transformation to handle long tail distributions \cite{zhao_ctab-gan_2021}. Through comparative evaluation against four other tabular data generators, it has been shown that \textit{CTABGAN} produces synthetic data with high utility, statistical similarity to real data, and reasonable safeguards for sensitive information \cite{zhao_ctab-gan_2021}. However, it is important to note that GAN-based models require a considerable number of samples for training, which should be considered when assessing the practical applicability of \textit{CTABGAN}.\par

    \textit{CTABGAN Plus} is an extended version of the \textit{CTABGAN} algorithm, designed to enhance the quality of synthetic data for machine learning utility and statistical similarity. This version introduces a new feature encoder tailored to variables following a single Gaussian distribution. It includes a redesigned shifted and scaled min-max transformation for normalizing these variables \cite{zhao_ctab-gan_2024}. Additionally, \textit{CTABGAN Plus} features a mixed-type encoder, which effectively represents mixed categorical-continuous variables and handles missing values. The algorithm utilizes the Wasserstein distance with a gradient penalty loss to improve the stability and effectiveness of GAN training \cite{weng_gan_2019}. Furthermore, an auxiliary classifier or regressor model is integrated to enhance synthesis performance for classification and regression tasks \cite{zhao_ctab-gan_2024}. \textit{CTABGAN Plus} introduces a newly designed conditional vector that uses log probabilities instead of original frequencies to address mode-collapse in imbalanced categorical and continuous variables \cite{zhao_ctab-gan_2024}. Moreover, it integrates efficient Differential Privacy (DP) into tabular GAN training through the DP-SGD algorithm, which trains a single discriminator, reducing complexity \cite{zhao_ctab-gan_2024}. The results show that \textit{CTABGAN Plus} produces synthetic data with higher machine-learning utility and greater similarity than ten baselines across seven tabular datasets \cite{zhao_ctab-gan_2024}.\par

    \textit{\textit{TVAE}} is a type of deep learning model that utilizes probabilistic modeling techniques to generate synthetic data \cite{vahdat_nvae_2020}. VAEs are equipped with both an encoder and a decoder, which allows them to introduce a unique approach to the latent space and inherent variability in the generated data. To maximize the likelihood of observed data and minimize the divergence between the latent distribution and a predefined prior, VAEs use a reparameterization strategy that enables backpropagation through the stochastic sampling process \cite{kingma_introduction_2019}. While VAEs can enhance data augmentation and enrichment, they face challenges in handling discrete data and may experience information loss, posterior collapse, and sensitivity to prior distribution \cite{pesteie_adaptive_2019}.\par
    
    \textit{TabDDPM} is a state-of-the-art generative model introduced by Kotelnikov \textit{et al.}\@ \cite{kotelnikov_tabddpm_2023} that utilizes diffusion models to generate tabular data \cite{zheng_diffusion_2023}. It applies noise to the input data through iterative diffusion and then reverses the process to create synthetic samples that closely resemble the original distribution \cite{kotelnikov_tabddpm_2023}. \textit{TabDDPM} addresses variations in feature types by preprocessing continuous features using Gaussian quantile transformation and representing categorical features with one hot encoding. The model employs multinomial diffusion for categorical features and Gaussian diffusion for continuous features, and each categorical feature undergoes a separate diffusion process \cite{kotelnikov_tabddpm_2023}. A multilayer perceptron predicts Gaussian and multinomial diffusion outcomes during the reverse diffusion step. The training process minimizes mean squared error for Gaussian diffusion and KL divergence for multinomial diffusion \cite{kotelnikov_tabddpm_2023}. \textit{TabDDPM} has demonstrated strong performance on benchmark datasets, consistently generating high-quality synthetic samples compared to existing GAN or VAE-based models. \par
    
    \textit{NextConvGeN} is designed to create synthetic tabular data using convex space learning. It generates synthetic samples similar to the original data by working within the boundaries of the original data neighborhoods \cite{mahendra_convex_2024}. The generator in \textit{NextConvGeN} combines batches of closely located real data points to learn the convex coefficients. This process is repeated iteratively between two neural networks to learn convex combinations. \textit{NextConvGeN} uses the same generator-discriminator architecture as ConvGeN \cite{schultz_convgen_2022} but is customized for tabular datasets. The generator operates on randomized neighborhoods of real data points rather than minority-class neighborhoods, which are determined using the Feature Distributed Clustering (FDC) approach. FDC effectively stratifies high-dimensional clinical tabular data \cite{bej_accounting_2023}. The discriminator in \textit{NextConvGeN} learns to classify synthetic points against shuffled batches of data points sampled from the complement of the input neighborhood provided to the generator, aiming to enhance classification performance.\par

    \textit{TabuLa}, developed by Zilong Zhao \textit{et al.}\@ \cite{zhao_tabula_2023} in 2023, is a new LLM-based framework to synthesize tabular data. The primary goal of \textit{TabuLa}
    is to accelerate the convergence speed of LLM-based methods for tabular data synthesis tasks. Two notable state-of-the-art tabular data synthesizers, GReaT \cite{borisov_language_2023} and REaLTabFormer \cite{solatorio_realtabformer_2023}, are based on LLMs but are hindered by extensive training time. \textit{TabuLa} addresses this challenge through four key features:
    i) Using a randomly initialized language model for data synthesis instead of relying on pre-trained models utilized in GReaT and REaLTabFormer. This strategic decision allows for a faster adaptation of the model to the requirements of tabular data synthesis tasks \cite{zhao_tabula_2023}. ii) Initialization of a foundational model from scratch and optimization specifically for tabular synthesis tasks, departing from the traditional reliance on pre-trained models \cite{zhao_tabula_2023}. This contributes to an effective learning process. iii) Reduction of token sequence length by consolidating all column names and categorical values into a single token each. This significantly decreases training time and enhances the model’s capability to efficiently learn and represent the relationships during training \cite{zhao_tabula_2023}. iv) Implementation of middle padding instead of the conventional left or right padding. This ensures that features within the same data column in the original data maintain identical absolute positions in the newly encoded token sequence, thereby enhancing the representation of tabular data for LLMs and resulting in improved synthesis quality \cite{zhao_tabula_2023}. The results of experiments using \textit{TabuLa} on seven different datasets demonstrate that \textit{TabuLa} reduces training time per epoch by an average of 46.2\% compared to the current state-of-the-art algorithm based on LLMs. Additionally, \textit{TabuLa} consistently achieves even higher utility with synthetic data \cite{zhao_tabula_2023}.\par
    Generative models discussed above have been evaluated based on the quality of the synthetic data, particularly in terms of utility and privacy measures. However, preserving inter-attribute dependencies in synthetic data has yet to be thoroughly explored. Preserving these dependencies is crucial for downstream analysis. The extent to which current models retain inter-attribute dependencies in the synthetic data remains an open question. Furthermore, there is no standard way to measure these dependencies. This study seeks to fill this gap by evaluating how well these models preserve functional and logical dependencies. It introduces a novel measure, the $Q$-function, to capture logical dependencies and utilizes the publicly accessible FDTool to identify functional dependencies. The detailed findings of this analysis are presented in Section \ref{sec: results}.
    

\section{Introducing the Q-function for quantifying inter-attribute logical dependencies} \label{Q_function}

Using the notation from the introduction, we have a tabular dataset $T$ with $m \geq 2$ columns and $n \geq 1$ rows.
The term $T_{i, j}$ denotes the value in the $i$-th row and $j$-th column.
The tuple $T_{i, \mathcal{A}} = (T_{i,\alpha_1}, T_{i, \alpha_2}, T_{i, \alpha_3}, \ldots)$ represents the values of row $i$ for the columns selected with the tuple $\mathcal{A} = (\alpha_1, \alpha_2, \alpha_3, \ldots)$.
The set $A = \{ T_{i, \mathcal{A}} \colon i = 1, 2, \ldots , n \}$ is the set of all tuples that exists in the table for the given selection of columns $\mathcal{A}$.
We defined the selection of columns $\mathcal{B}$ and the set of tuples $B = \{ T_{i, \mathcal{B}} \colon i = 1, 2, \ldots , n \}$ for that selection in the same way.

Define the relation $\sim_T \subseteq A \times B$ between $A$ and $B$ with $a \sim_T b$ holds if and only if $a \in A$ and $b \in B$ are in the same row in the table $T$ (meaning there is an $i$ so that $T_{i, \mathcal{A}} = a$ and $T_{i. \mathcal{B}} = b$).

Define $C^* = \{ (\alpha_1, \ldots ) \colon \{ \alpha_1, \ldots \} \subseteq \{ 1, 2, 3, \ldots , m\}\}$ as the set containing all possible selection of columns.

The aim was to have a function $Q_T : C^* \times C^* \to [0,1]$ that:
\begin{itemize}
\item $Q_T(\mathcal{A}, \mathcal{B})$ should be zero if the table and column selections produces a function $f : A \to B$
\item $Q_T(\mathcal{A}, \mathcal{B})$ should be one if for all $a \in A$ and all $b \in B$ the relation $a \sim_T b$ holds.
\item Every other value of $Q_T(\mathcal{A}, \mathcal{B})$ should be between zero and one
\end{itemize}

We construct this function as follows:
For a given $a \in A$ we can count how many $b \in B$ holds $a \sim_T b$ by $|D(a)| = |\{ b \in B \colon a \sim_T b \}|$.
The more values are excluded, the closer we are to a function that takes this $a$ and gives us a $b$.
If there is a function from $A$ to $B$ for this value $a$, then $|D(a)| = 1$.
By the definition of our table (not empty) and $A, B$ (= sets of the actual values in the table) we know that $|D(a)|$ has to be greater or equal to 1. By subtracting 1, we get: $n' = |D(a)| - 1 \geq 0$

If $a \sim_T b$ for all $b\in B$ for this $a \in A$ then $n' = |B| - 1$. Assuming $|B| > 1$ we can divide this value and get the function:
\[
  G_B(a) = \frac{|D(a)| - 1}{|B| - 1}
\]
that maps to the interval $[0, 1]$. We can collect this information for all $a \in A$ by 
\[
  s_0 = \sum_{a \in A} G_B(a)
  = \sum_{a \in A} \frac{|D(a)| - 1}{|B| - 1}
  = \frac{1}{|B| - 1} \sum_{a \in A} \big(|F(a)| - 1\big)
\]
We can easily see that $0 \leq s_0 \leq |A|$. Dividing $|A|$ gives us one solution to our wishlist for $|B| > 1$:
\[
  s_1 = \frac{s_0}{|A|}
  = \frac{1}{|A| \cdot (|B| - 1)} \sum_{a \in A} \big(|D(a)| - 1\big)
  = \frac{1}{|A| \cdot (|B| - 1)} \sum_{a \in A} \Big(\left|\big\{ b \in B \colon a \sim_T b \big\}\right| - 1\Big)
\]
If we put $a$ in the set, we can omit the sum and get the easier term:
\[
  s_1 = \frac{\left|\big\{ (a,b) \colon a \in A, b \in B \mbox{ and } a \sim_T b \big\}\right| - |A|}{|A| \cdot (|B| - 1)}
\]
For the special case that there is only one value in column $B$ (e.g. $B = \{ b \}$) we have the obvious constant function $f(a) = b$.
In the special case of an empty set $A$ or empty set, $B$, there is no pair left to disprove the existence of a function.
This leads to our final definition:
\begin{equation}
\begin{array}{lll}
 Q_T(\mathcal{A}, \mathcal{B} ) & = \frac{|\{ (a,b) \colon a \in A, b \in B \text{ and } a \sim_T b \}| - |A|}{|A| \cdot (|B| - 1)} &  \text{if } |A| \geq 1 \text{ and }|B| > 1 \\
 Q_T(\mathcal{A}, \mathcal{B})   & = 0  &  \text{if } |A| = 0 \text{ or } |B| \leq 1 
\end{array}
\label{q-function}
\end{equation}


\section{Experimental protocols}

Preserving logical and functional dependencies is essential for assessing the quality of synthetic tabular data. To address this, we conducted a comparative study to investigate how well different generative models preserve these dependencies. The experimental workflow in Figure \ref{workflow} involved generating synthetic tabular data using \textit{CTGAN}, \textit{CTABGAN}, \textit{CTABGAN Plus}, \textit{\textit{TVAE}}, \textit{NextConvGeN}, \textit{TabDDPM}, and \textit{TabuLa}. The following are the steps involved in the experiment:\par
\begin{itemize}
    \item \textbf{Data preparation and model training:} We used seven well-known generative models for effectively generating synthetic tabular data. Unlike the usual practice of splitting data into training and testing subsets, we trained the models on the complete tabular dataset. This approach ensured that the models were fully exposed to the nuances of the data, potentially improving their ability to replicate the underlying inter-attribute relationships. We used the default parameters to train all models.
    \item \textbf{Generation of synthetic data:} After training each generative model, we generated synthetic data of the same size as the original dataset, ensuring equal comparisons between real and synthetic data and facilitating a more accurate evaluation of the models' performance.
    \item \textbf{Extraction of functional and logical dependencies:} We utilized the FDTool algorithm \cite{buranosky_fdtool_2019} to extract functional dependencies from real and synthetic datasets, focusing on categorical features. We employed the $Q$-function approach outlined in this paper to evaluate logical dependencies. Following this, we graphed the Q-scores for the real and synthetic data and identified feature pairs with similar scores in both datasets, revealing logical dependence.
    \item \textbf{Comparative analysis:} We used Venn diagrams to compare the functional dependencies in real and synthetic datasets. These diagrams show where the dependencies overlap, clearly showing how well the generative models maintain functional relationships between the attributes (Figure \ref{Airbnb_FD} and \ref{Migraine_FD}. Additionally, we used bar plots to illustrate the percentage of logical dependencies preserved by the synthetic data compared to the real data. This visual representation makes it easier to understand how each model performs (Figure \ref{LD comparison}).

\end{itemize}

\subsection{Datasets used in the comparative analysis:}
There are no well-known datasets that have functional and logical dependencies. Finding datasets with such conditions is difficult. We have chosen five publicly available datasets for our bench-marking analysis. The choice of datasets covers a variety of situations, including observational studies, clinical trials, and surveys; therefore, it is more comprehensive. There is also variation in the number of attributes they contain, ranging from one to $11$ continuous attributes and from eight up to $17$ categorical attributes, and the size of datasets varies from $377$ to $4908$. Table \ref{dataset_table} describes the dataset used in the experiment. 

\begin{table*}[!ht]
\footnotesize
\begin{center}
 \begin{tabular}{c|c|c|c|c|c} 
  \hline
  
   {\textbf{Dataset}}&  {\textbf{Sample size}} &  {\textbf{Attributes}} &  {\textbf{Continuous}} & {\textbf{Categorical}} & {\textbf{Target variable}}\\
   \hline
   
   \href{https://codeocean.com/capsule/1269964/tree/v1/data/migraine.csv}{Migraine} & $377$ & $20$ & $1$ & $17$ & Type \\
   \hline
  
   \href{https://www.kaggle.com/datasets/fedesoriano/cirrhosis-prediction-dataset/code}{Liver cirrhosis} & $418$ & $19$ & $11$ & $8$ & Stage  \\
   \hline

   \href{https://doi.org/10.24432/C5H31Z}{Obesity} & $2087$ & $17$ & $3$ & $14$ & NObeyesdad \\
   \hline

   \href{https://www.kaggle.com/competitions/airbnb-recruiting-new-user-bookings/data}{Airbnb} & $3048$ & $18$ & $4$ & $14$ & country destination \\
   \hline 

   \href{https://www.kaggle.com/datasets/fedesoriano/stroke-prediction-dataset/code}{Stroke} & $4908$ & $11$ & $3$ & $8$ & stroke \\
   \hline
   
\end{tabular}
\caption{Description of the datasets used in the experiment. This table provides information on the dataset size, number of attributes, distribution of continuous and categorical features, and the target variable within each dataset.}
\label{dataset_table}
\end{center}
\end{table*}

\section{Results}\label{sec: results} Our experiment is designed to assess the ability of tabular generative models to retain inter-attribute logical and functional relationships from real data to synthetic data using the Q-function and FDTool algorithms.\par
\begin{figure}[ht]
    \centering
    \includegraphics[width=\textwidth]{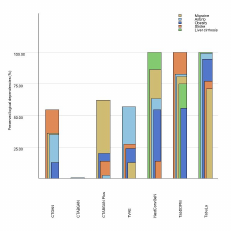}
    \caption{The chart illustrates the preservation of logical dependencies across different generative models, with the x-axis representing the models and the y-axis indicating the percentage of preserved logical dependencies. Each color corresponds to a different dataset. \textit{NextConvGeN}, \textit{TabDDPM}, and \textit{TabuLa} models consistently exhibit higher percentages for all datasets, demonstrating their ability to retain more logical dependencies compared to other models.}

    \label{LD comparison}
\end{figure}

 
\textbf{GAN-based and VAE models fail to capture logical dependencies in the data:} For features $f_i$ and $f_j$ in a tabular data $T$, the smaller the value of
$|Q_T(\{f_i\},\{f_j\})-Q_S(\{f_i\},\{f_j\})|$ the better the functional and logical dependencies are preserved for a synthetic table $S$.
We investigated this with various generative models and noticed that \textit{CTGAN}, \textit{CTABGAN}, \textit{CTABGAN Plus}, and \textit{\textit{TVAE}} do not always effectively maintain logical connections compared to other generative models (refer to Figure \ref{LD comparison}). One explanation is that training GANs requires a large amount of data, and our experiments were conducted using small tabular datasets, which is more consistent with real-world scenarios from the biomedical domain.\par

We observed that, among the GAN models, \textit{CTGAN} showed comparatively better performance in preserving logical dependencies in the Stroke dataset due to its relatively larger size ($4908$ samples). From Figure \ref{Stroke_Q-metric}, we can deduce that the \textit{CTGAN}-generated synthetic Stroke data produces a similar $Q$-score compared to real Stroke data for more feature pairs, as we observe that there are more points along the diagonal line of the plot. On the other hand, \textit{CTABGAN} failed to maintain a single logical dependency, except for the Airbnb dataset, which had a preservation rate of only $0.68$\%. This lack of preservation can be observed in Figures \ref{migrain_Q-metric}, \ref{Obesity_Q-metric}, \ref{Stroke_Q-metric}, \ref{Liver-cirrhosis_Q-metric}, where all the points lie at one, indicating no dependencies between feature pairs. Our results indicate that the \textit{CTABGAN} model focuses on modeling multimode and long-tail distributions rather than preserving dependencies.\par

Interestingly, the \textit{CTABGAN Plus} model, despite having fewer data points, exhibited a higher percentage of logical dependency preservation in the Migraine dataset compared to other datasets. Note that the Migraine dataset contains $17$ categorical attributes, the highest among all the datasets used in the experiment. The \textit{CTABGAN Plus} model incorporates a specialized encoder to effectively handle mixed categorical attributes, making it adept at capturing relationships and distributions within categorical attributes, contributing to its superior performance in logical dependency preservation.\par

The \textit{\textit{TVAE}} could capture logical dependencies only for the Airbnb dataset, for which it achieved a preservation rate of around $56.79$\%. The main reason behind this limitation of \textit{\textit{TVAE}} is the phenomenon of mode collapse. \textit{\textit{TVAE}}s compress the data into a latent space where each dimension captures some aspect of the data. However, the dominant category may occupy a larger portion of the latent space for imbalanced attributes. This could cause the model to generate the same value for the entire column, which is not ideal for analyzing logical dependencies between attributes. Therefore, GANs and \textit{TVAE}s may not be the best choice for certain datasets where logical dependencies between attributes are important and require accurate modeling.\par

\textbf{Convex space, diffusion-based, and transformer-based models are effective in preserving logical dependencies:} 
A direct illustrative comparison of \textit{NextConvGeN}, \textit{TabDDPM}, and \textit{TabuLa} models can be seen in Figure \ref{LD comparison}. Our experiments imply that the \textit{NextConvGeN} model performs well in Liver Cirrhosis and Migraine, with logical dependencies captured up to $86.39$\% and $100$\%, respectively. In comparison, \textit{TabDDPM} retains $75$\% and $80.92$\% of the logical dependencies in the Liver Cirrhosis and Migraine datasets and all the dependencies in the Stroke dataset. On the other hand, \textit{TabuLa} can preserve most of the inter-attribute logical dependencies present in real data for all datasets. Our results indicate that \textit{TabuLa}, \textit{TabDDPM}, and \textit{NextConvGeN} produce comparable performances in the context of logical dependency preservation, and these three models are superior to the rest of the compared models.\par

\begin{figure}[ht]
    \centering
    \begin{tabular}{|c|c|}
    \hline 
        \subfigure[CTGAN]{
            \includegraphics[width=0.35\linewidth]{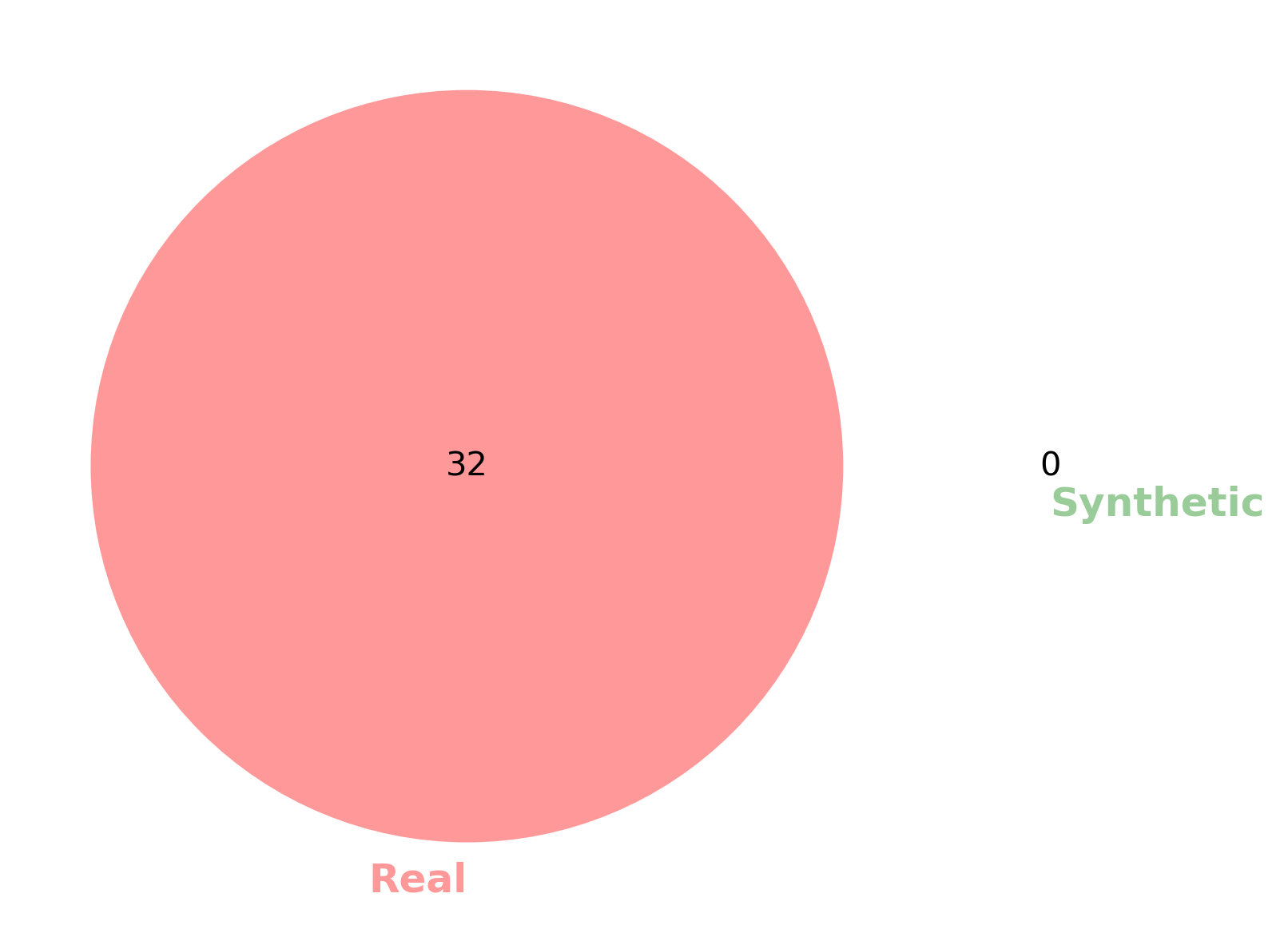}
        }
        &
        \subfigure[CTABGAN]{
            \includegraphics[width=0.35\linewidth]{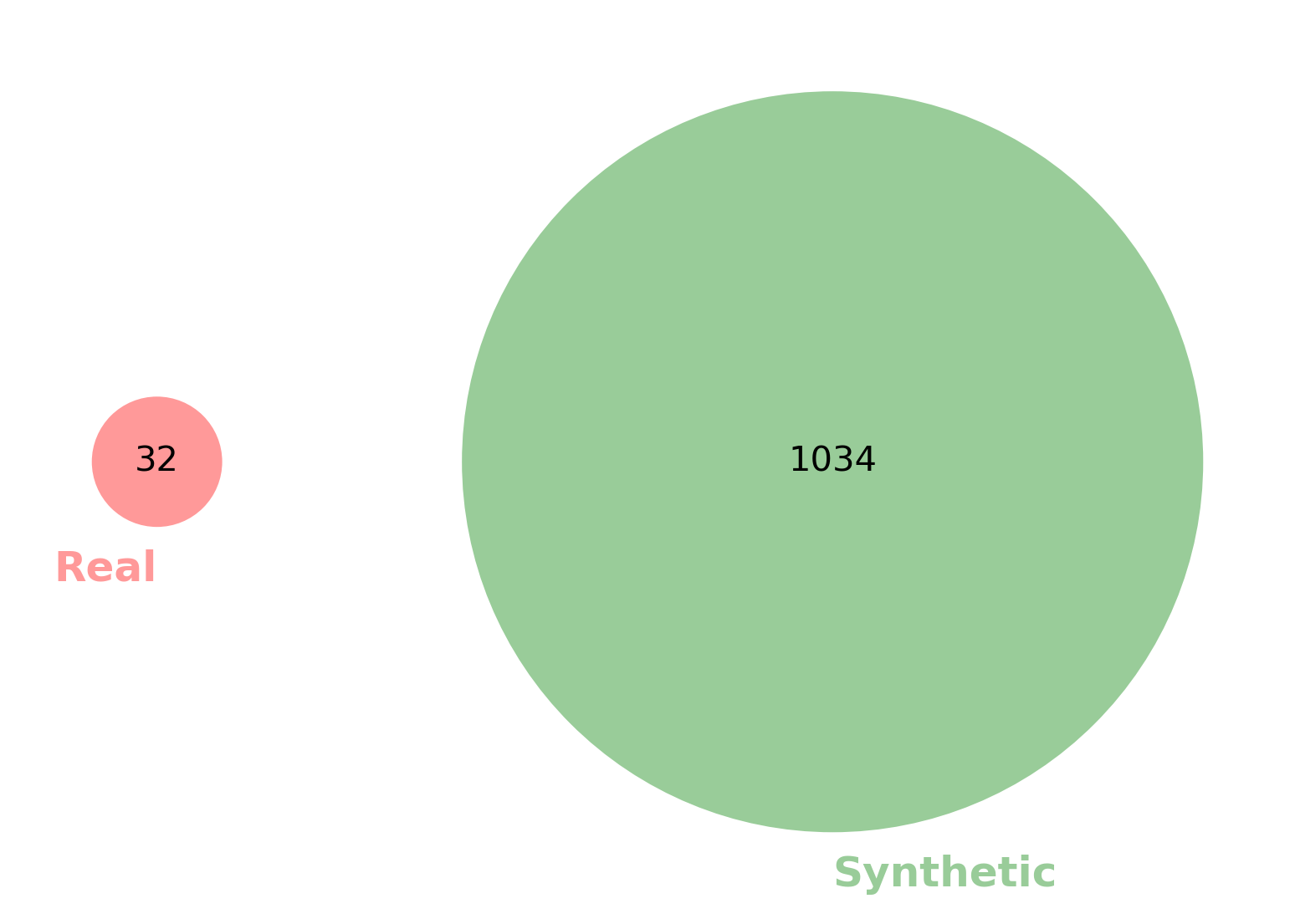}
        } \\
        \hline
        \subfigure[CTABGAN Plus]{
            \includegraphics[width=0.35\linewidth]{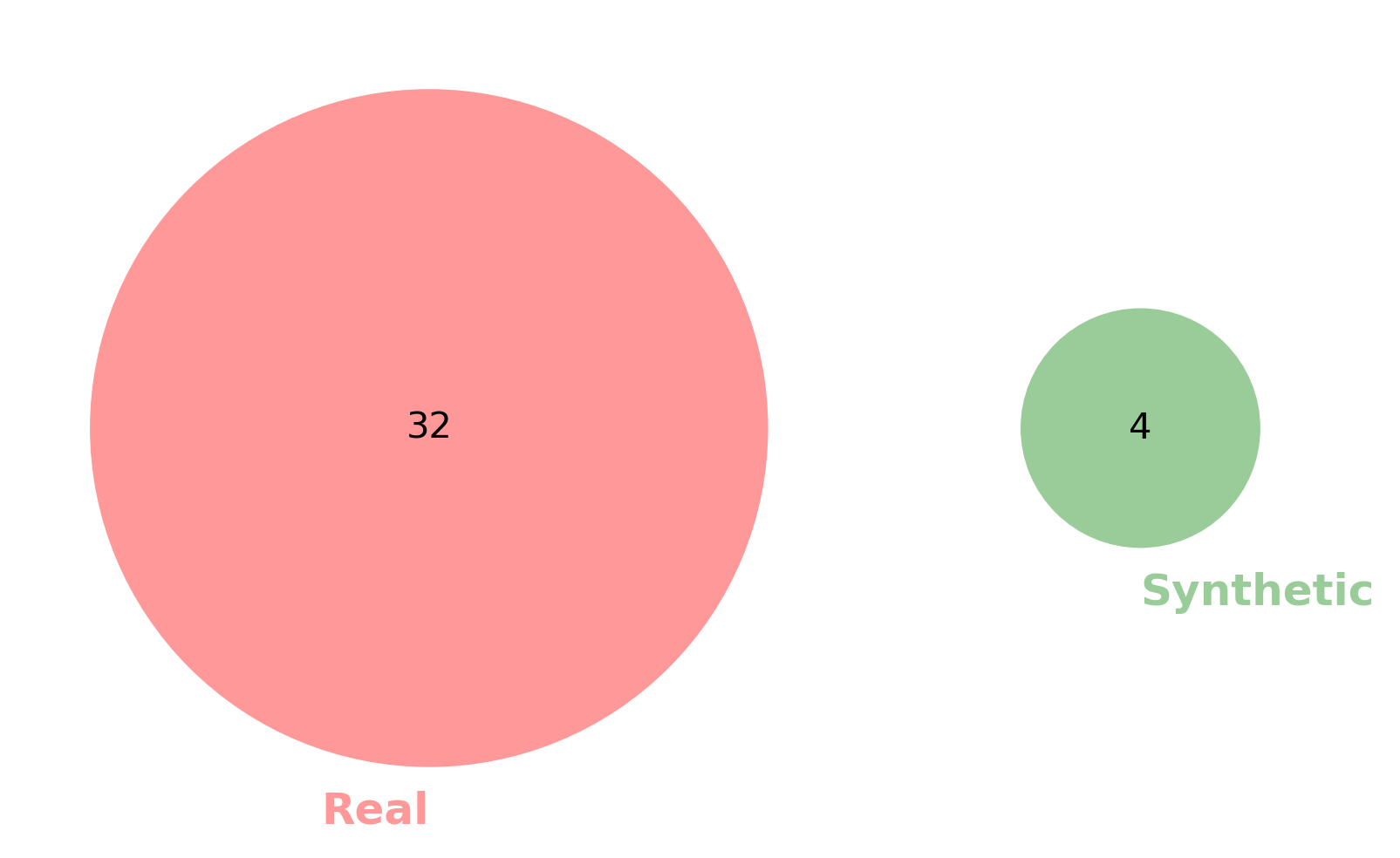}
        }
        &
        \subfigure[TVAE]{
            \includegraphics[width=0.35\linewidth]{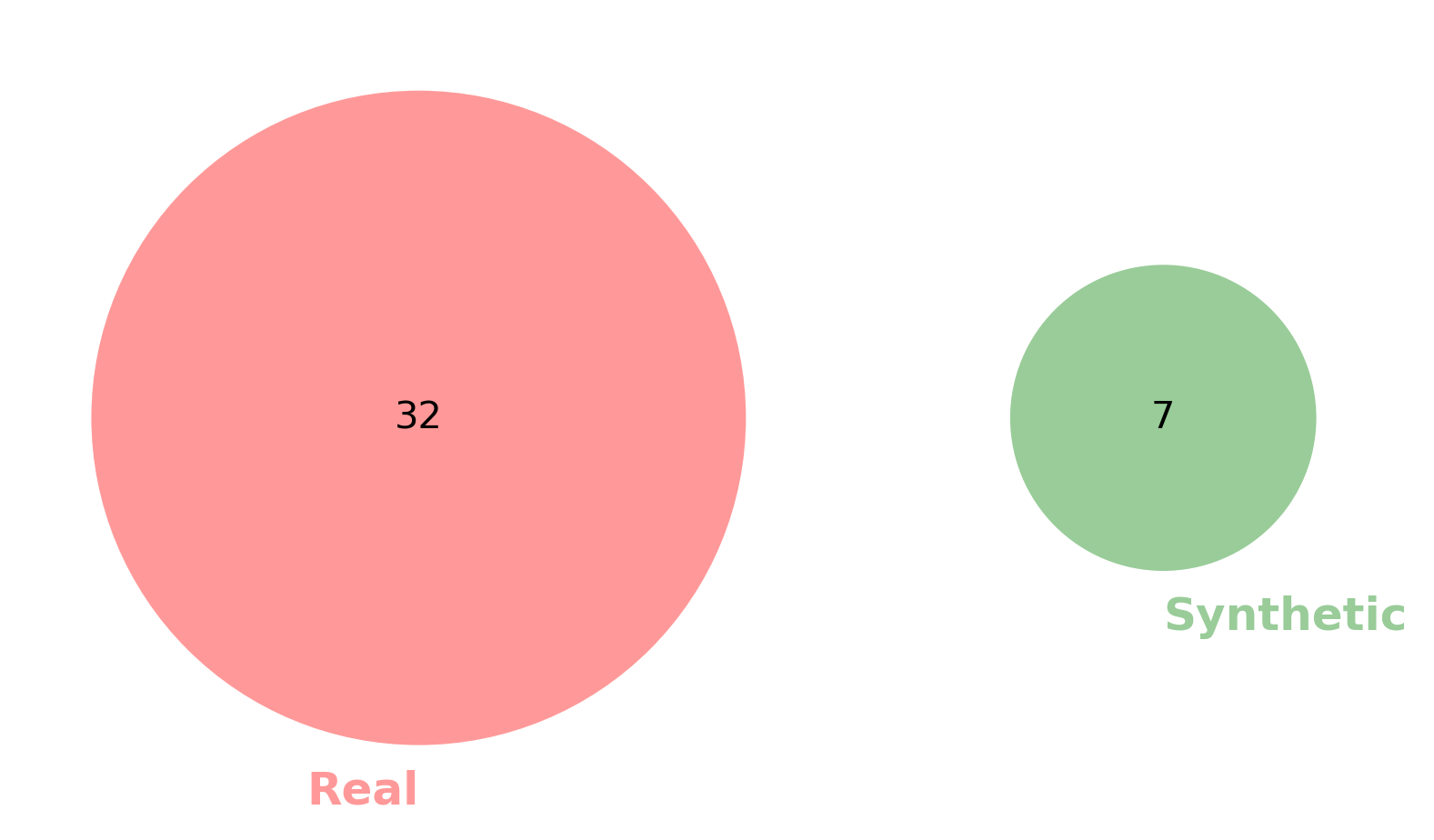}
        } \\
        \hline
        \subfigure[NextConvGeN]{
            \includegraphics[width=0.35\linewidth]{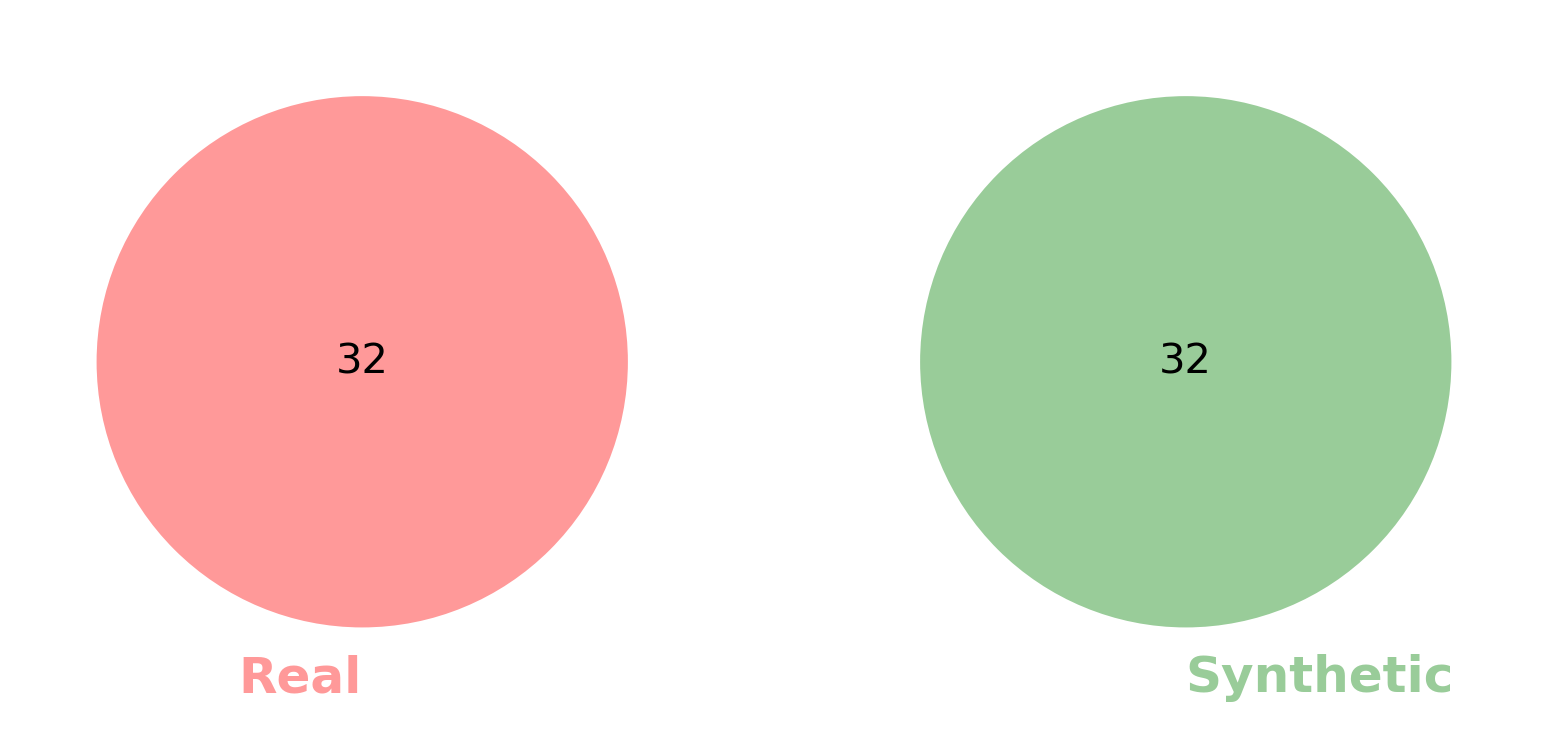}
        }
        &
        \subfigure[TabDDPM]{
            \includegraphics[width=0.35\linewidth]{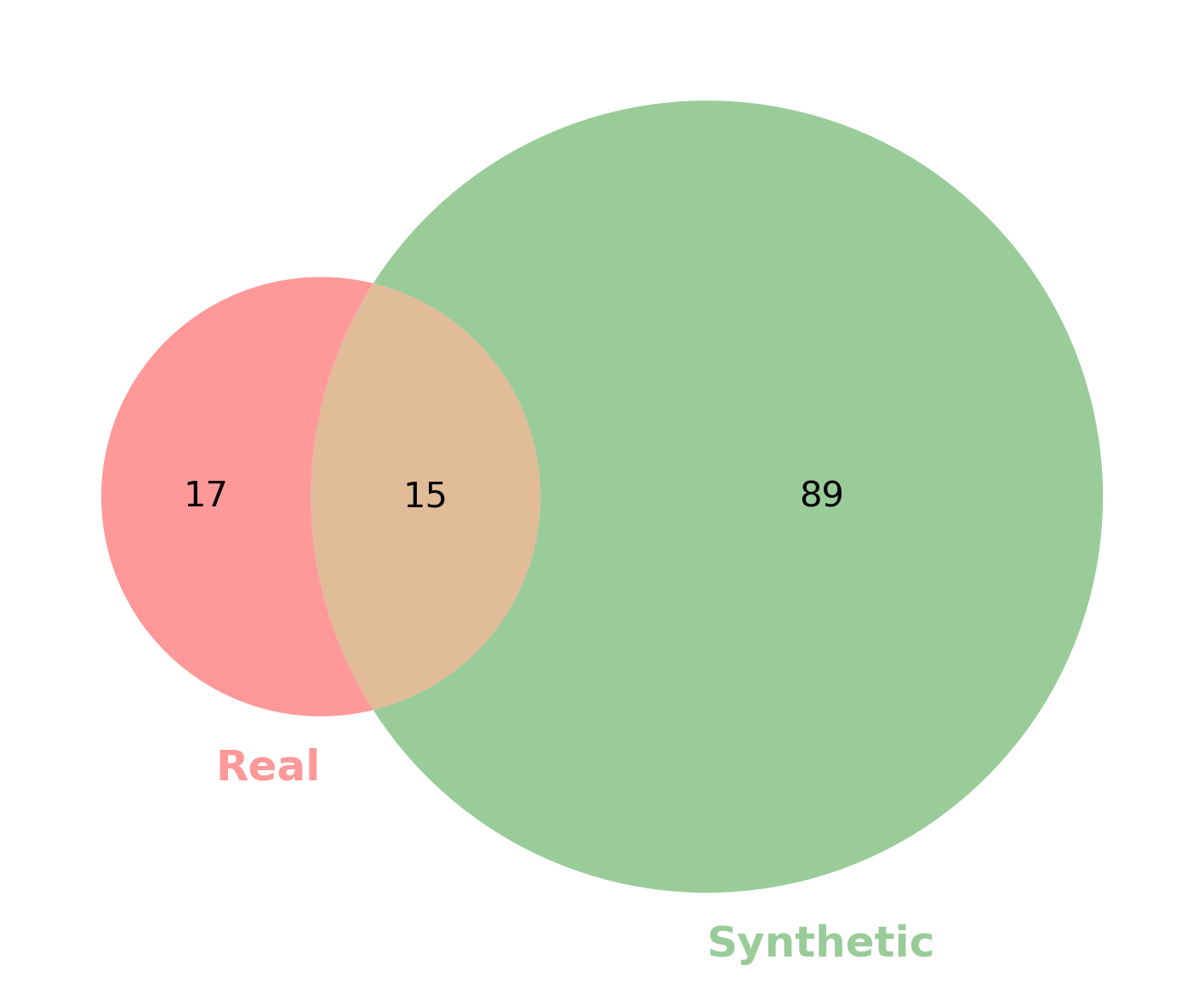}
        } \\
        \hline
        \multicolumn{2}{|c|}{
            \subfigure[TabuLa]{
                \includegraphics[width=0.35\linewidth]{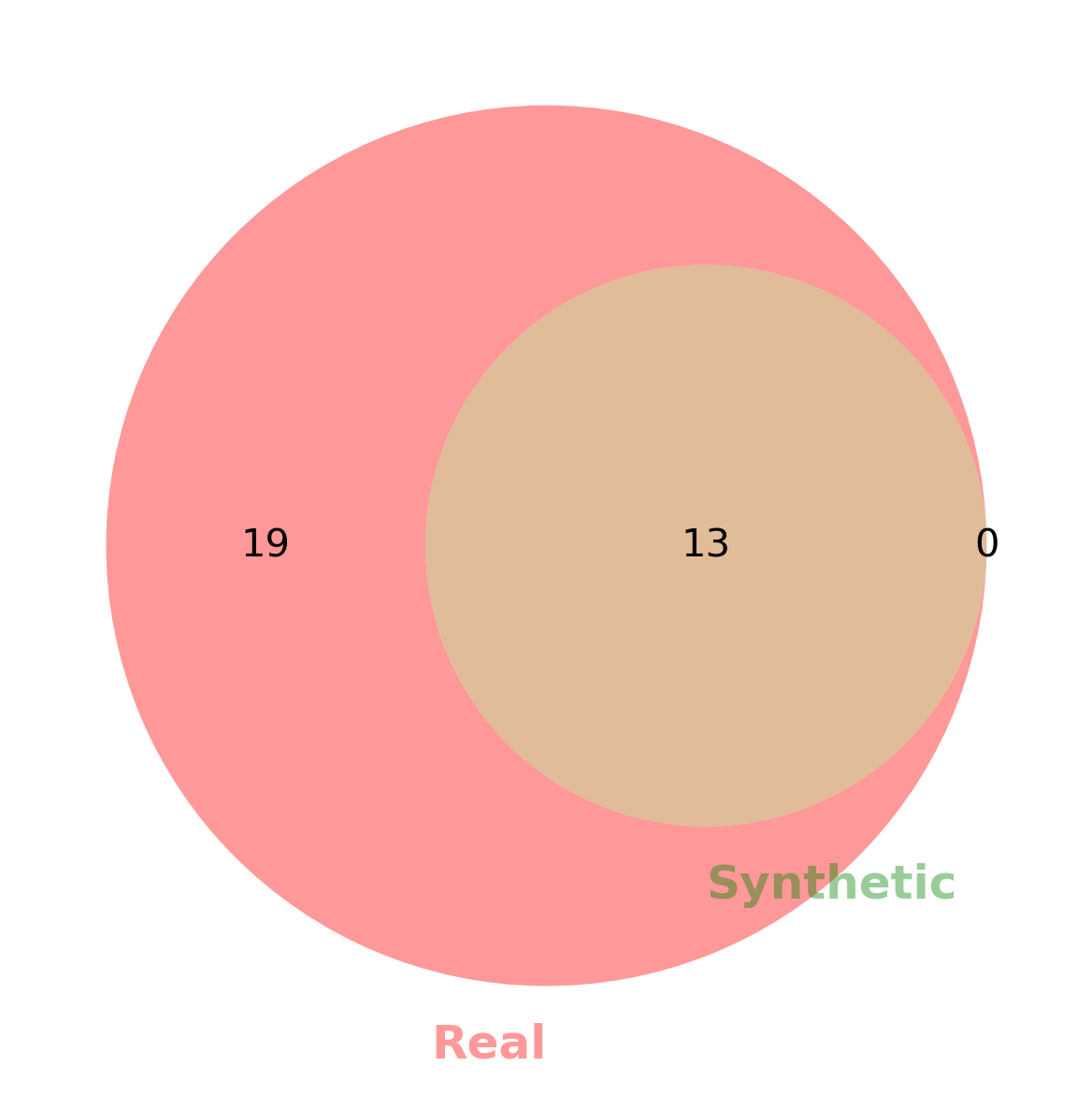}
            }
        } \\
        \hline
    \end{tabular}
    \caption{\textbf{Comparison of functional dependencies in Airbnb data:} The figure displays Venn diagrams comparing functional dependencies in real (coral) and synthetic (green) Airbnb data from seven generative models. Numbers within circles indicate total counts of dependencies. Overlap shows shared dependencies retained by synthetic data. Notably, none of the generative models manage to preserve a larger number of dependencies than the real data. However, \textit{TabDDPM} and \textit{TabuLa} succeed in preserving some functional dependencies.}
    \label{Airbnb_FD}
\end{figure}

\begin{figure}
    \centering
    \begin{tabular}{|c|c|}
    \hline 
        \subfigure[CTGAN]{
            \includegraphics[width=0.35\linewidth]{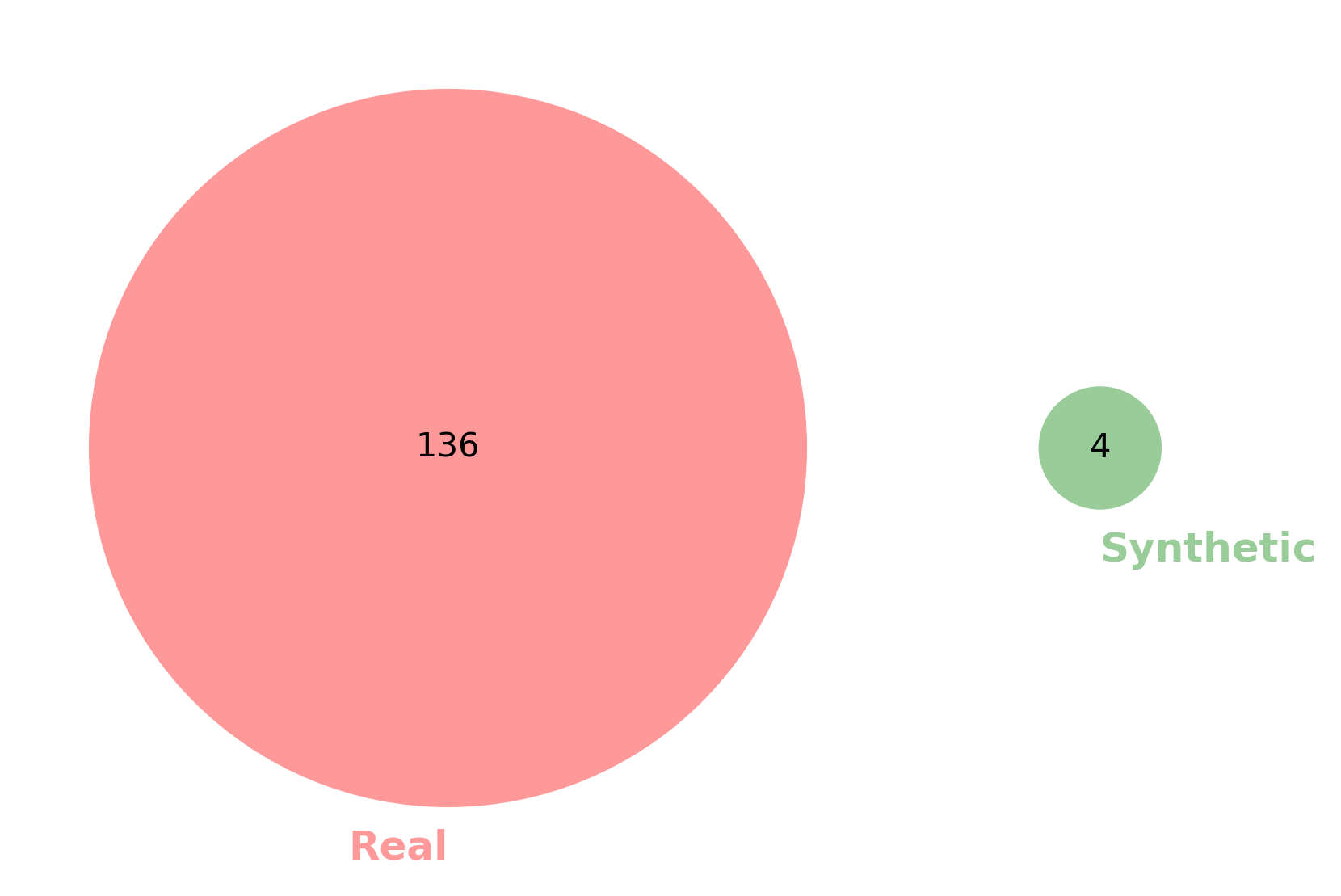}
            \label{fig:ven1}
        }
        &
        \subfigure[CTABGAN]{
            \includegraphics[width=0.35\linewidth]{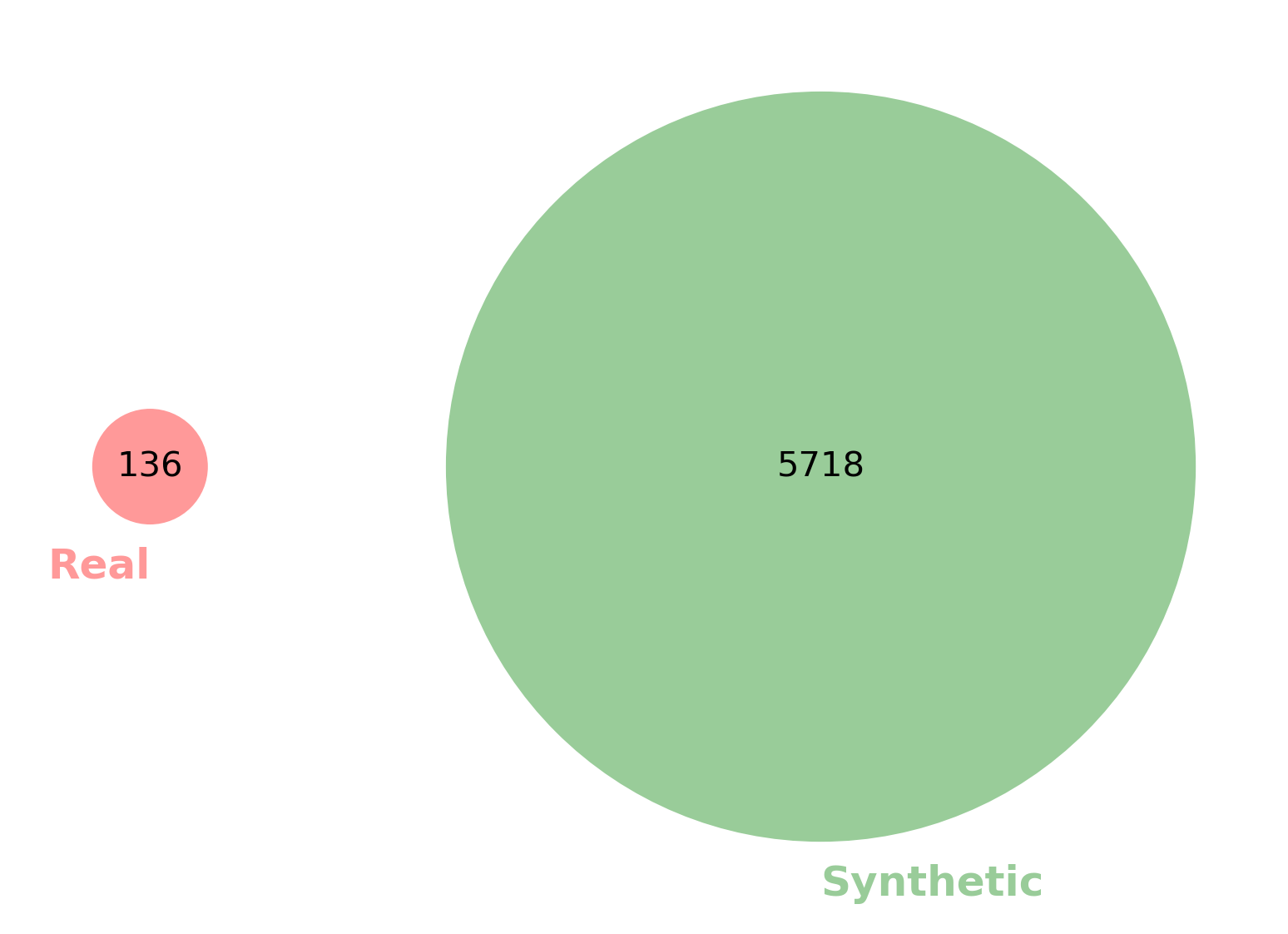}
            \label{fig:ven2}
        } \\
        \hline
        \subfigure[CTABGAN Plus]{
            \includegraphics[width=0.35\linewidth]{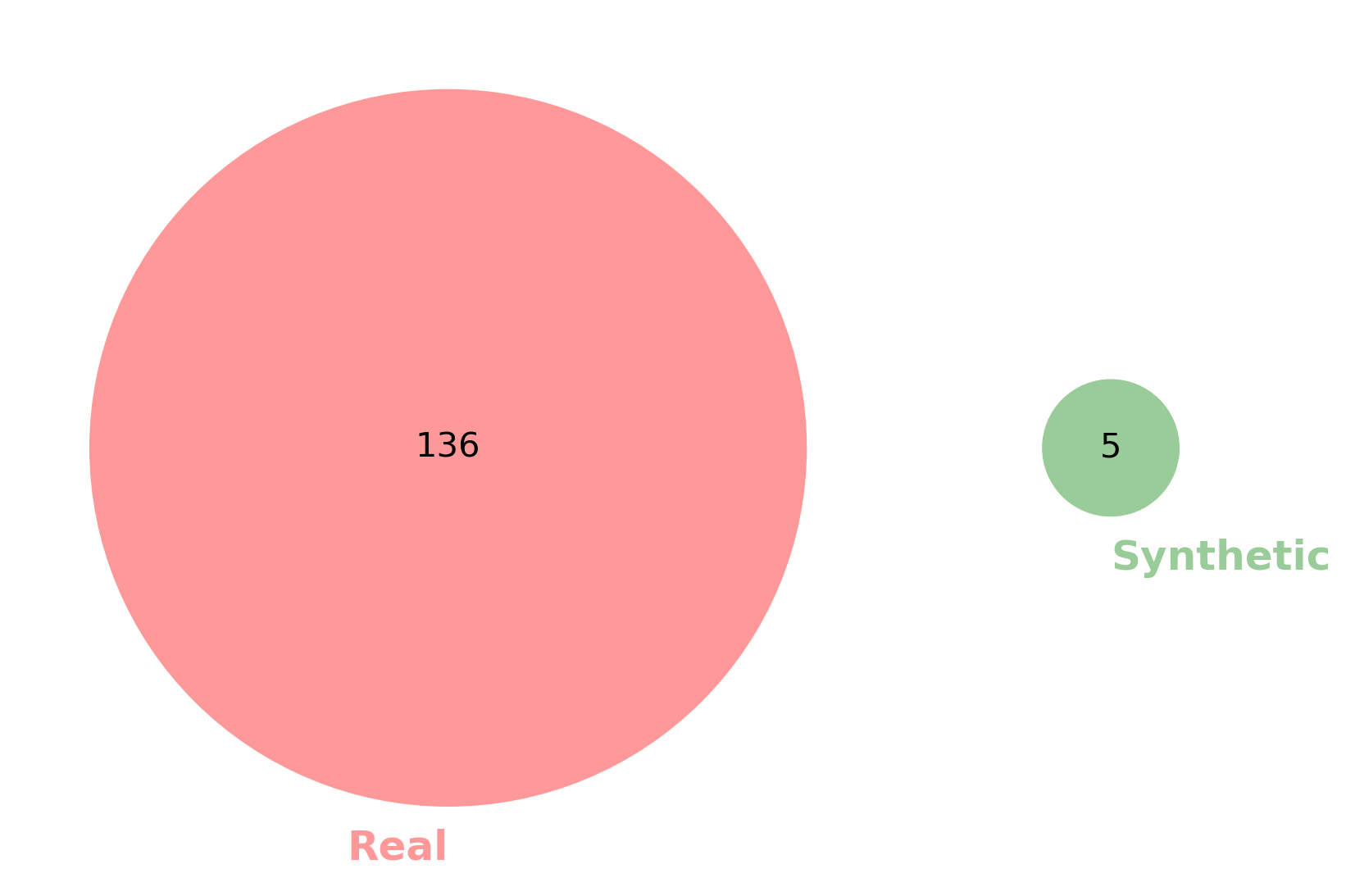}
            \label{fig:ven3}
        }
        &
        \subfigure[TVAE]{
            \includegraphics[width=0.35\linewidth]{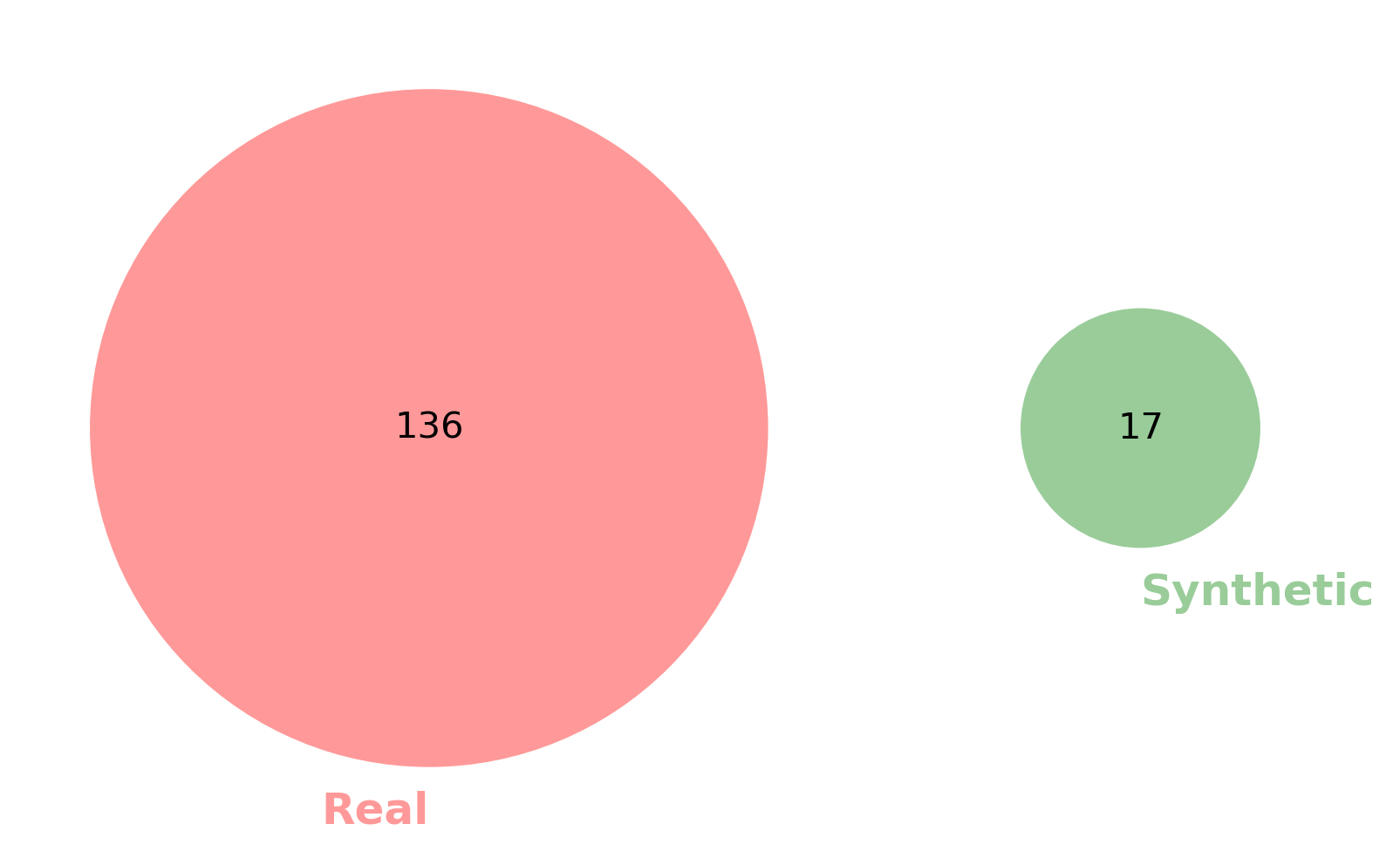}
            \label{fig:ven4}
        } \\
        \hline
        \subfigure[NextConvGeN]{
            \includegraphics[width=0.35\linewidth]{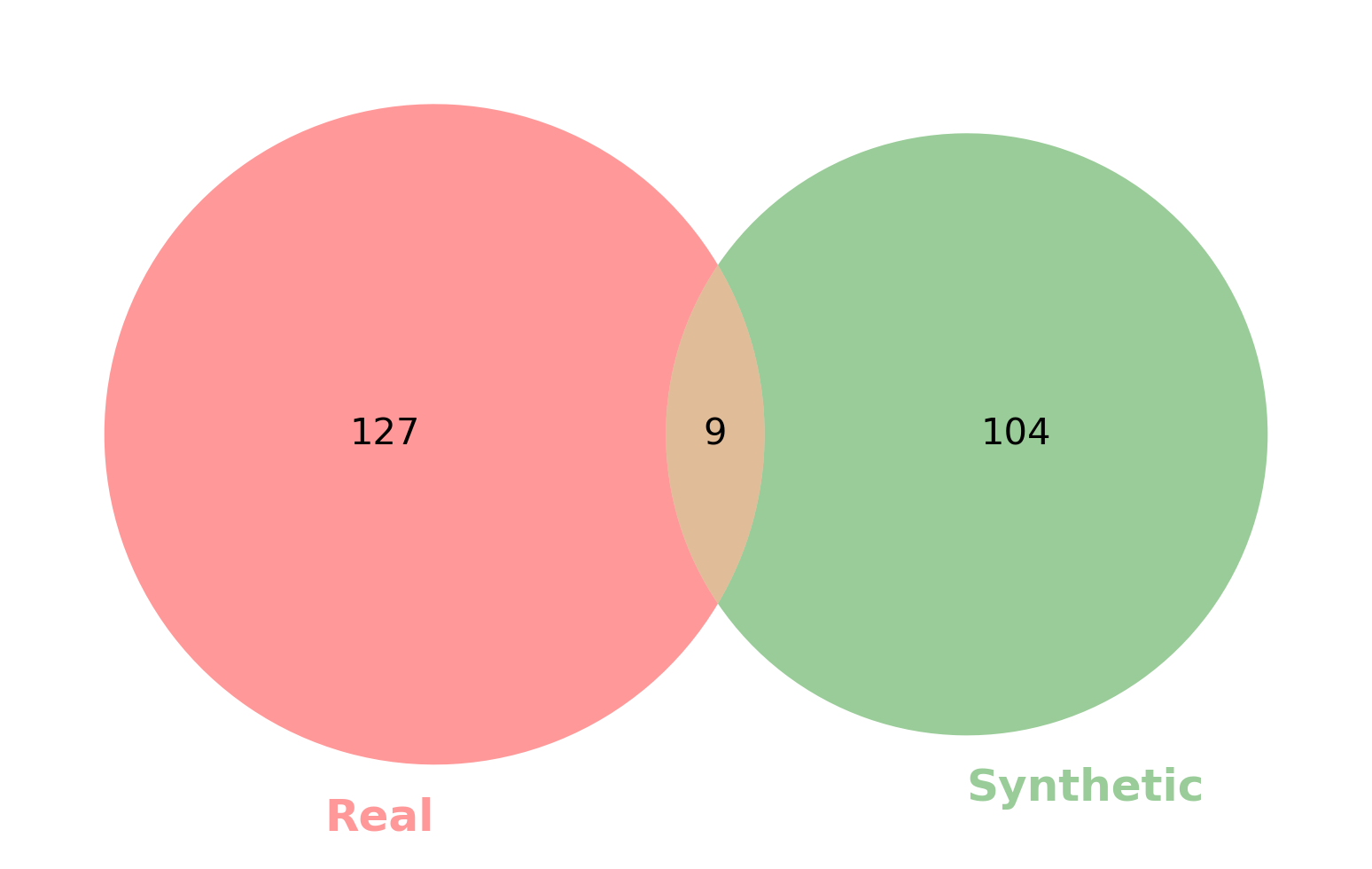}
            \label{fig:ven5}
        }
        &
        \subfigure[TabDDPM]{
            \includegraphics[width=0.35\linewidth]{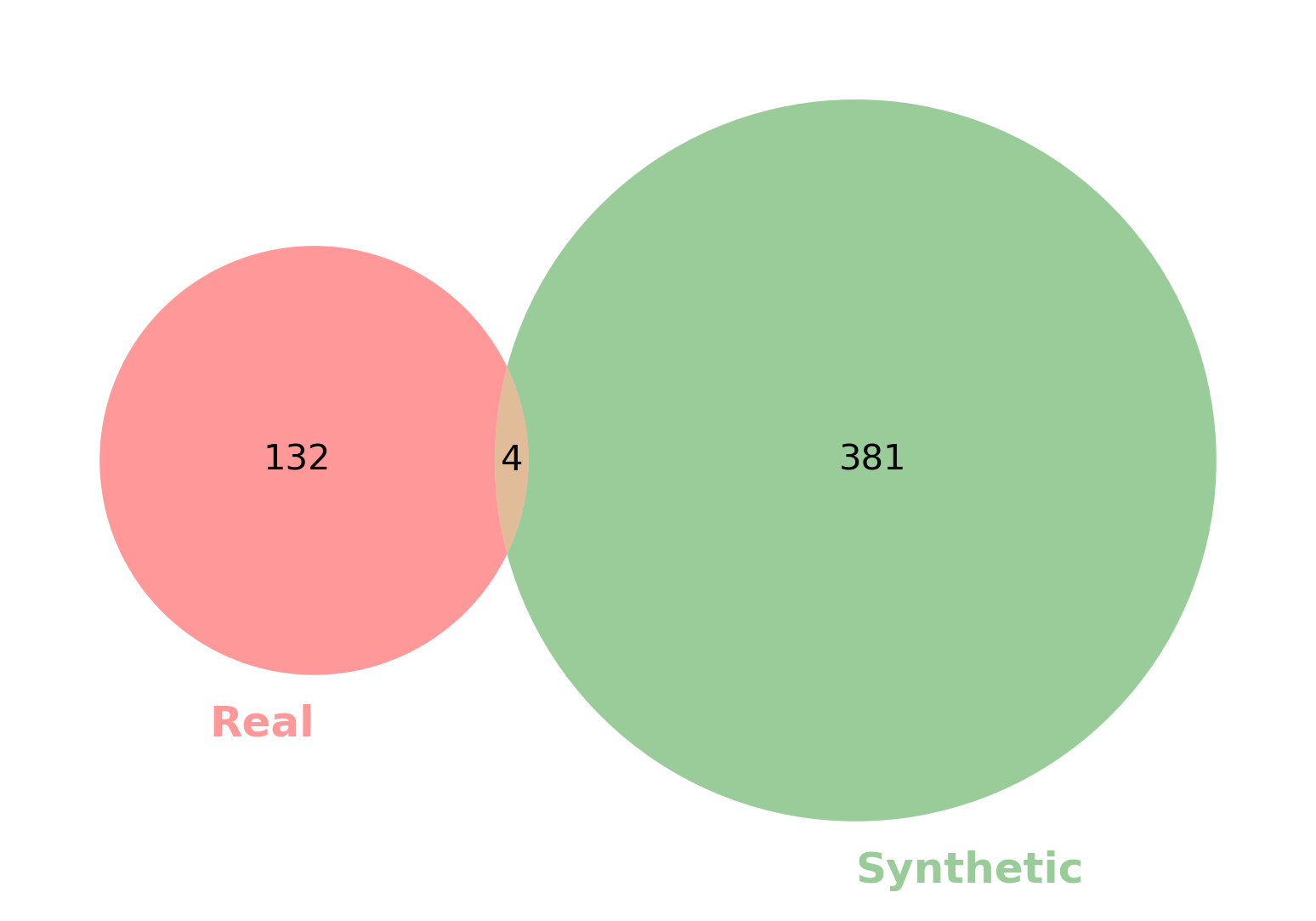}
            \label{fig:ven6}
        } \\
        \hline
        \multicolumn{2}{|c|}{
            \subfigure[TabuLa]{
                \includegraphics[width=0.35\linewidth]{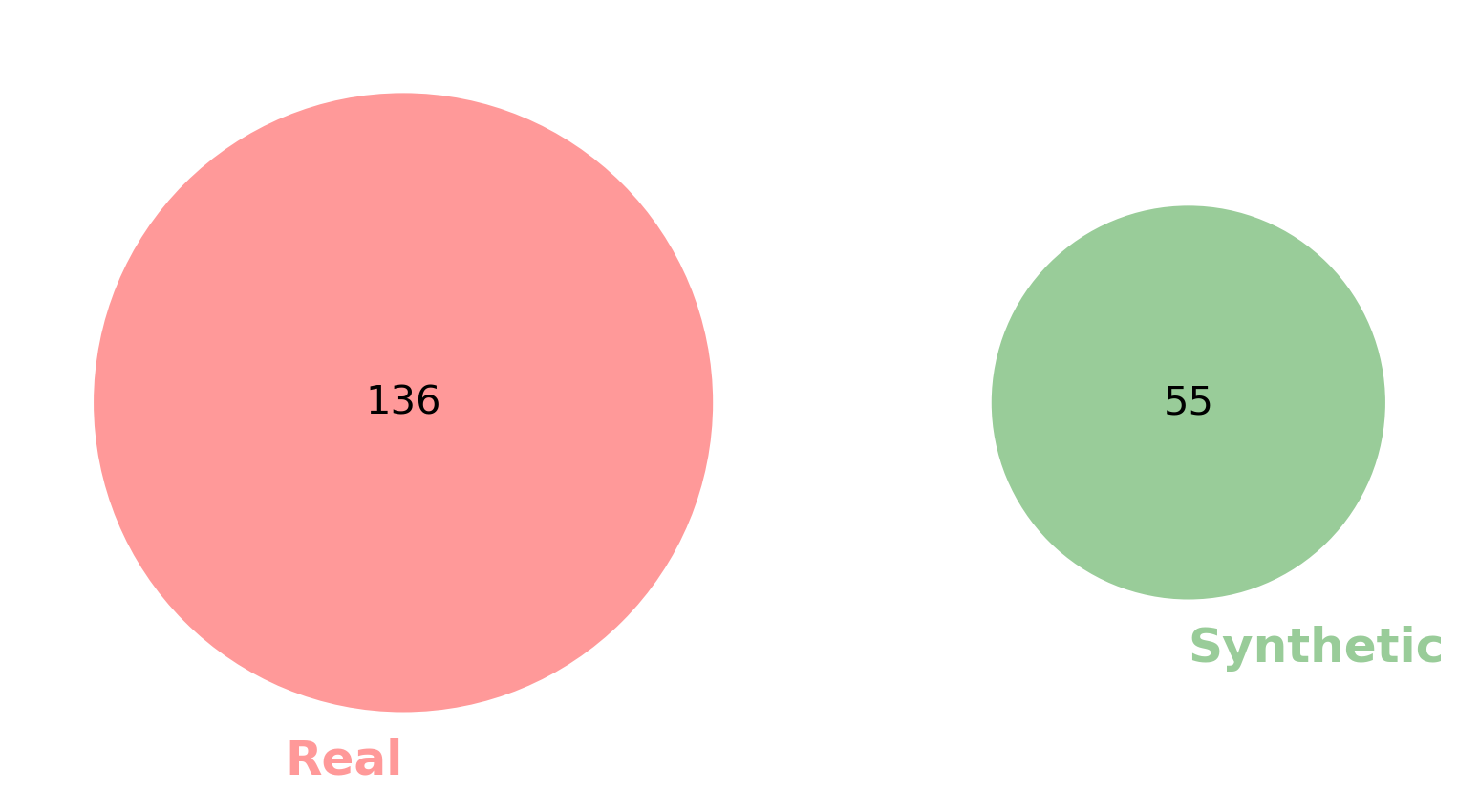}
                \label{fig:ven7}
            }
        } \\
        \hline
    \end{tabular}
    \caption{\textbf{Comparison of functional dependencies in Migraine data:} The figure displays Venn diagrams comparing functional dependencies in real (coral) and synthetic (green) Migraine data from various generative models. Numbers within circles indicate total counts of dependencies. Overlap shows shared dependencies retained by synthetic data. Notably, none of the generative models manage to preserve a larger number of dependencies than the real data. However, \textit{NextConvGeN} and \textit{TabDDPM} succeed in preserving some functional dependencies.}
    \label{Migraine_FD}
\end{figure}

\textbf{None of the generative models satisfactorily preserved inter-attribute functional dependencies:} Two datasets that exhibit functional dependencies are Airbnb and Migraine. The Migraine dataset displays $136$ functional dependencies, while the Airbnb dataset has $32$ functional dependencies. Tabular generative models tested on these datasets demonstrate that they can rarely capture the functional dependencies present in these datasets (See Figure \ref{Migraine_FD} and \ref{Airbnb_FD}). In particular, \textit{CTGAN}, \textit{CTABGAN Plus}, and \textit{TVAE} could not capture functional dependencies in the two mentioned datasets. Interestingly, in the synthetic data generated by \textit{CTABGAN}, functional dependencies were observed that were not present in the real data. \textit{NextConvGeN} managed to preserve only nine out of the $136$ functional dependencies in the Migraine dataset and none in the Airbnb dataset. \textit{TabDDPM} preserved four functional dependencies in the Migraine dataset and $17$ in the Airbnb dataset, but it also resulted in several functional dependencies that were not present in the actual data. \textit{TabuLa} preserved $13$ functional dependencies in the Airbnb dataset and none in the Migraine dataset. These results highlight the inability of current generative models to capture functional dependencies, particularly in datasets with a higher number of categorical features. While these models aim to generate synthetic data that resembles real data properties without compromising privacy, they were not designed to preserve functional dependencies. However, we would also like to point out that preserving functional dependencies in synthetic data can be challenging, as even a single contradiction in a data point in the synthetic data disregards two attributes in the synthetic dataset to be functionally dependent. In conclusion, observed that, unlike logical dependencies, current tabular generative models struggle to retain functional dependencies in synthetic data.\par

\clearpage

\clearpage

\section{Discussion}
One intriguing question that arises from our experiments is why GAN and VAE-based models are not as effective in preserving inter-attribute logical relationships as convex-space, diffusion, and transformer-based models. One key distinction is that GAN-based models rely on a generative approach that does not directly utilize real data to create synthetic samples. Instead, these models aim to map random noise to the distribution of the training data. This poses challenges when working with limited data, often encountered in several practical scenarios, such as clinical data obtained from a single healthcare facility, as substantial training data is required.

\begin{figure}[ht]
    \centering
    \includegraphics[width=\textwidth]{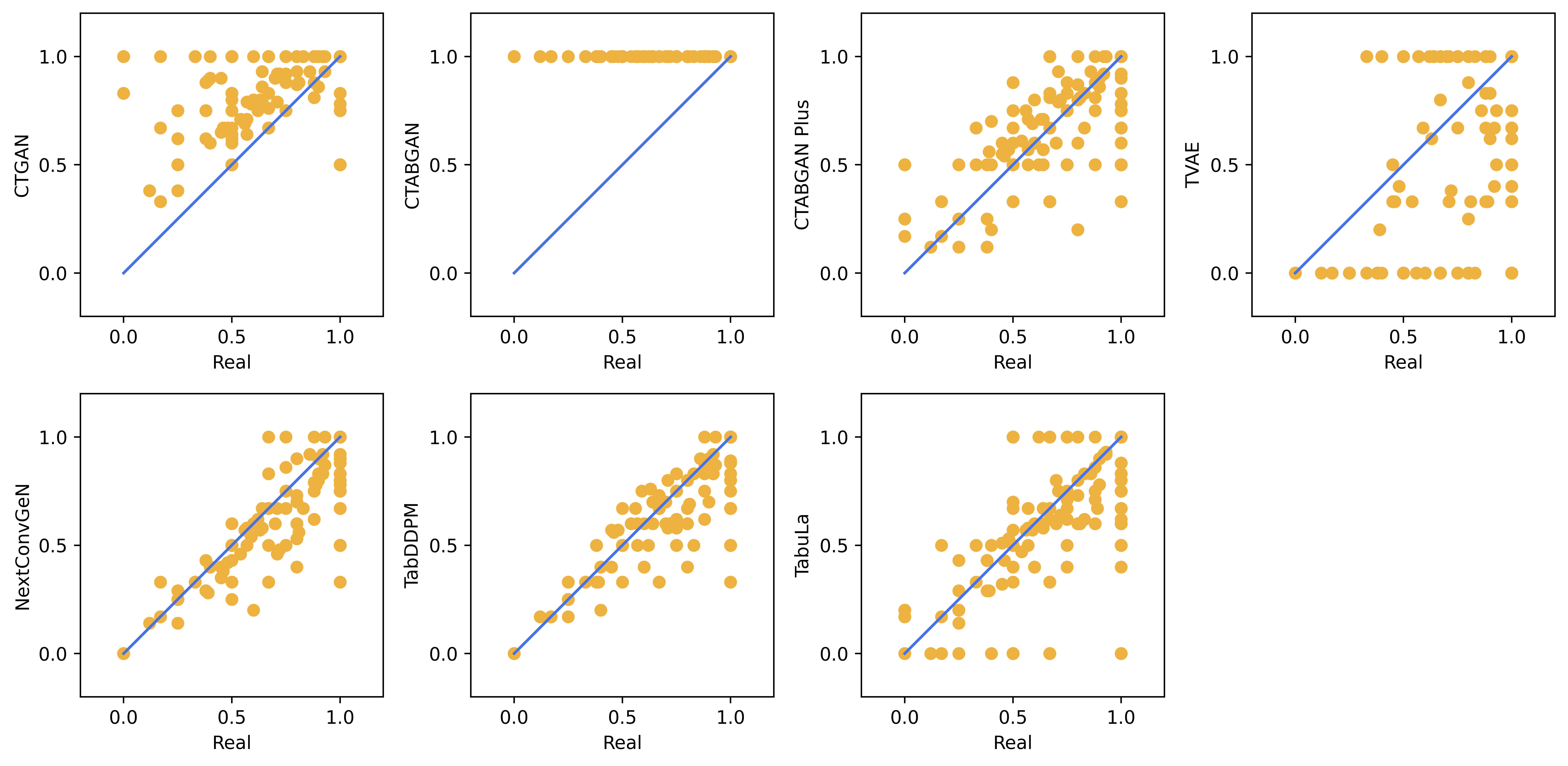}
    \caption{The plots show how closely the $Q$-scores of synthetic Migraine data match those of real Migraine data. Points on the diagonal line signify preserved dependencies, and the points above suggest some real dependencies that are not in synthetic data, while the points below imply that synthetic data introduces new dependencies. Generally, when more points are closer to or on the line, synthetic data better preserves dependencies than real data. \textit{NextConvGeN}, \textit{TabDDPM}, and \textit{TabuLa} effectively capture the logical dependencies in synthetic data, closely reflecting those present in the real data.}
    \label{migrain_Q-metric}
\end{figure}

\begin{figure}[ht]
    \centering
    \includegraphics[width=\textwidth]{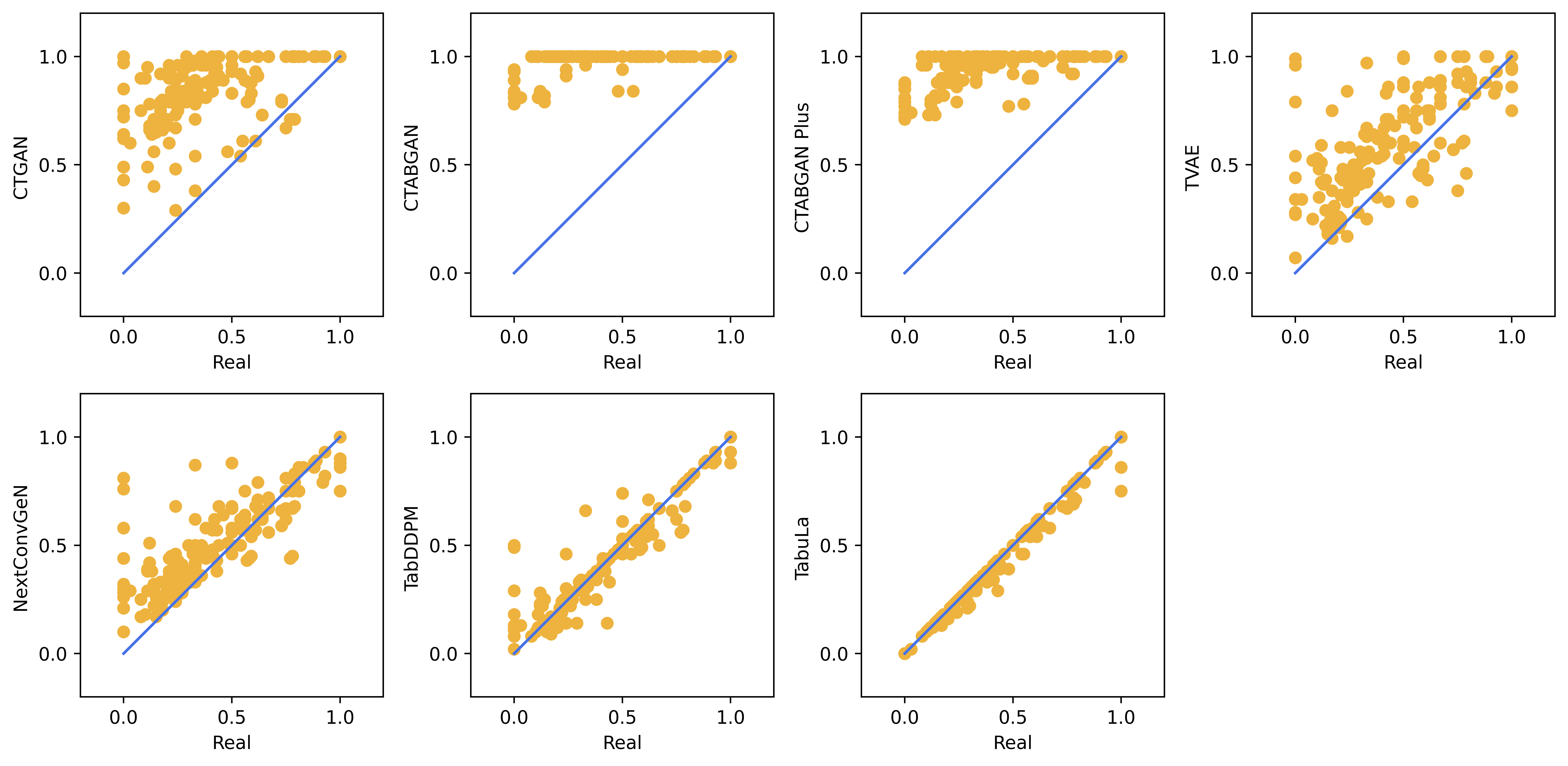}
    \caption{The plots show how closely the $Q$-scores of synthetic Airbnb data match those of real Airbnb data. Points on the diagonal line signify preserved dependencies, and the points above suggest some real dependencies that are not in synthetic data, while the points below imply that synthetic data introduces new dependencies. Generally, when more points are closer to or on the line, synthetic data better preserves dependencies than real data. \textit{NextConvGeN}, \textit{TabDDPM}, and \textit{TabuLa} effectively capture the logical dependencies in synthetic data, closely reflecting those present in the real data.}
    
    \label{airbnb_Q-metric}
\end{figure}

\begin{figure}[ht]
    \centering
    \includegraphics[width=\textwidth]{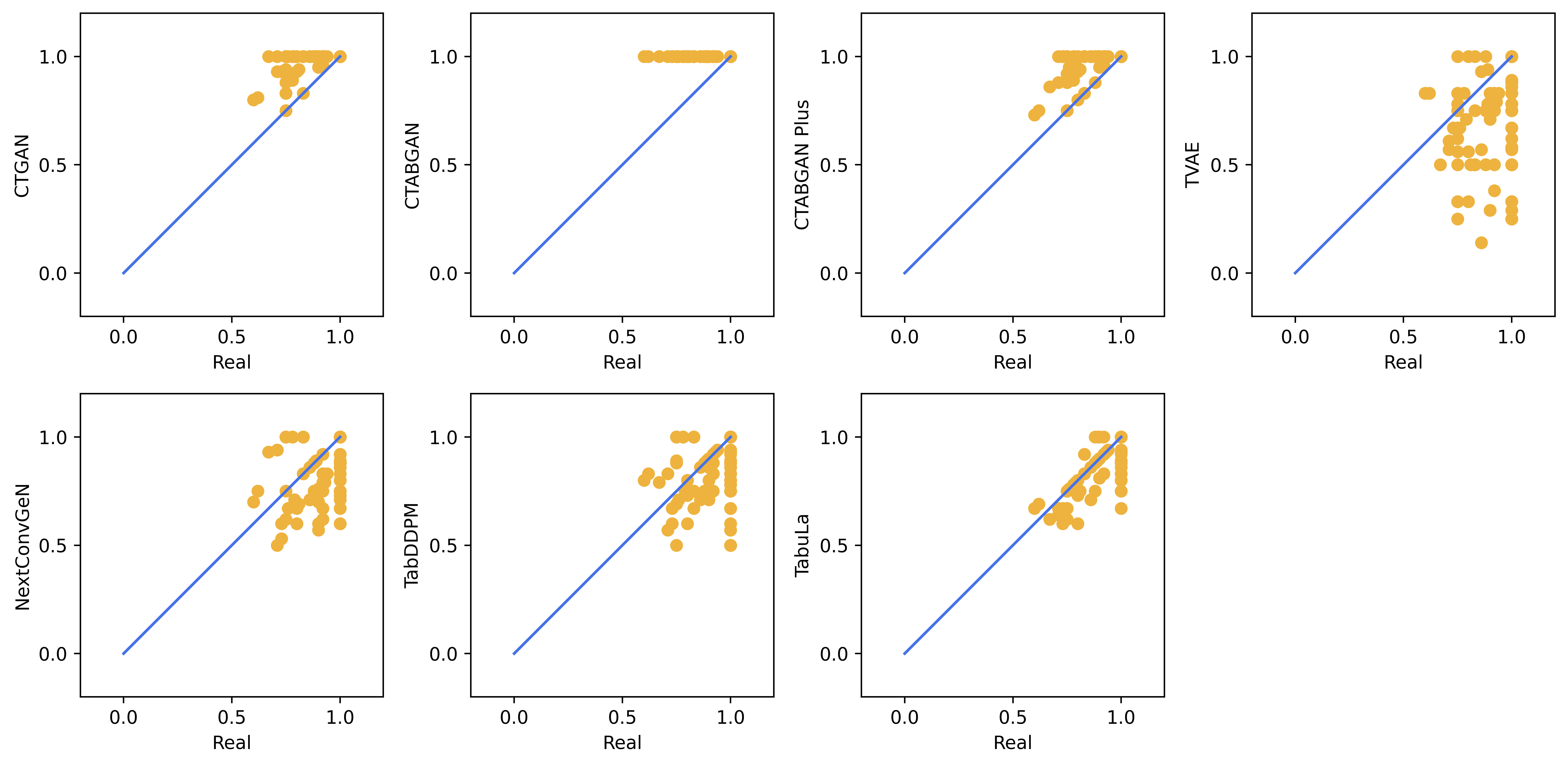}
    \caption{The plots show how closely the $Q$-scores of synthetic Obesity data match those of real Obesity data. Points on the diagonal line signify preserved dependencies, and the points above suggest some real dependencies that are not in synthetic data, while the points below imply that synthetic data introduces new dependencies. Generally, when more points are closer to or on the line, synthetic data better preserves dependencies than real data.}
    
    \label{Obesity_Q-metric}
\end{figure}

\begin{figure}[ht]
    \centering
    \includegraphics[width=\textwidth]{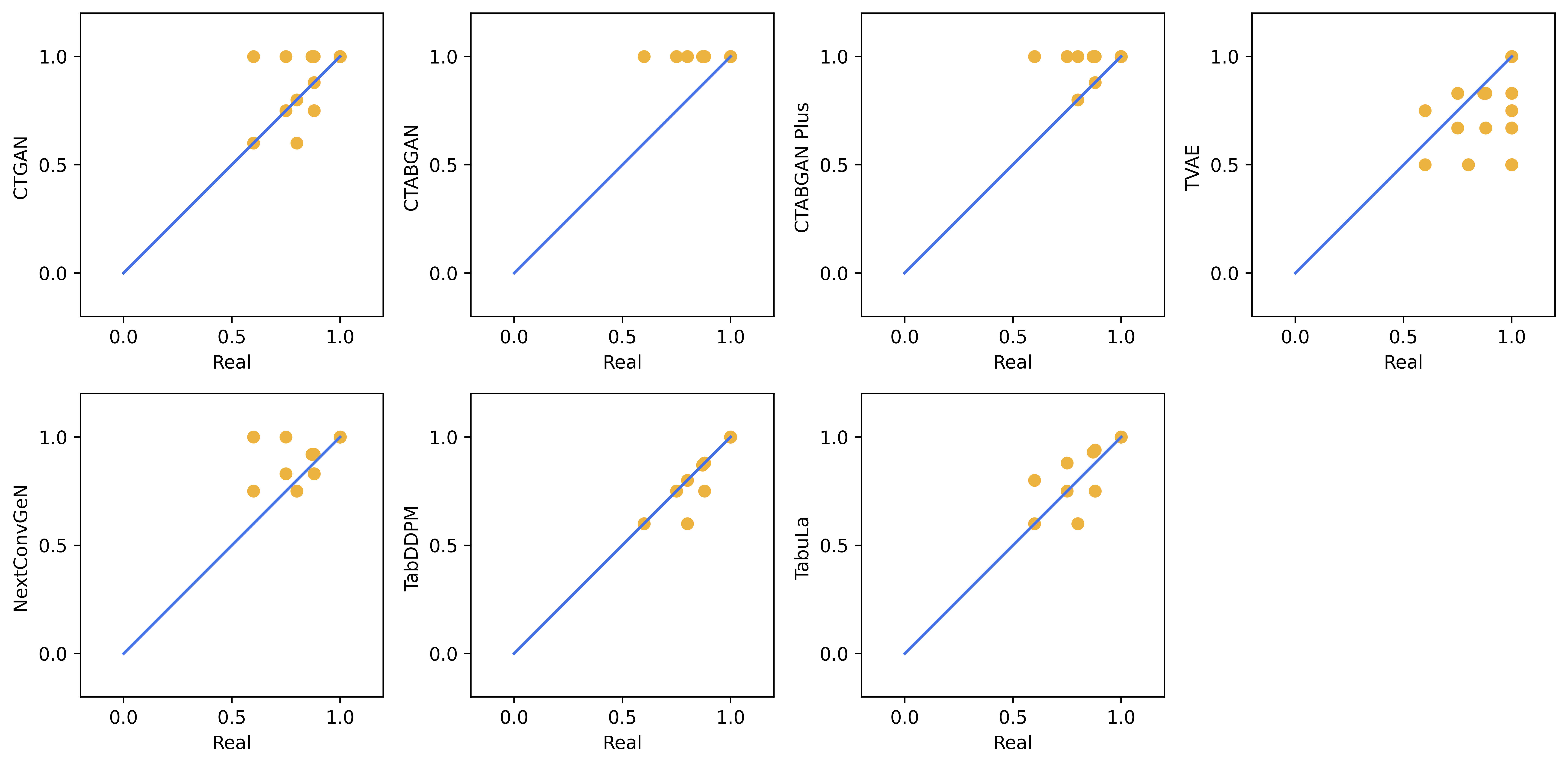}
    \caption{The plots show how closely the $Q$-scores of synthetic Stroke data match those of real Stroke data. Points on the diagonal line signify preserved dependencies, and the points above suggest some real dependencies that are not in synthetic data, while the points below imply that synthetic data introduces new dependencies. Generally, when more points are closer to or on the line, synthetic data better preserves dependencies than real data. \textit{NextConvGeN}, \textit{TabDDPM}, and \textit{TabuLa} effectively capture the logical dependencies in synthetic data, closely reflecting those present in the real data.}
    
    \label{Stroke_Q-metric}
\end{figure}

\begin{figure}[ht]
    \centering
    \includegraphics[width=\textwidth]{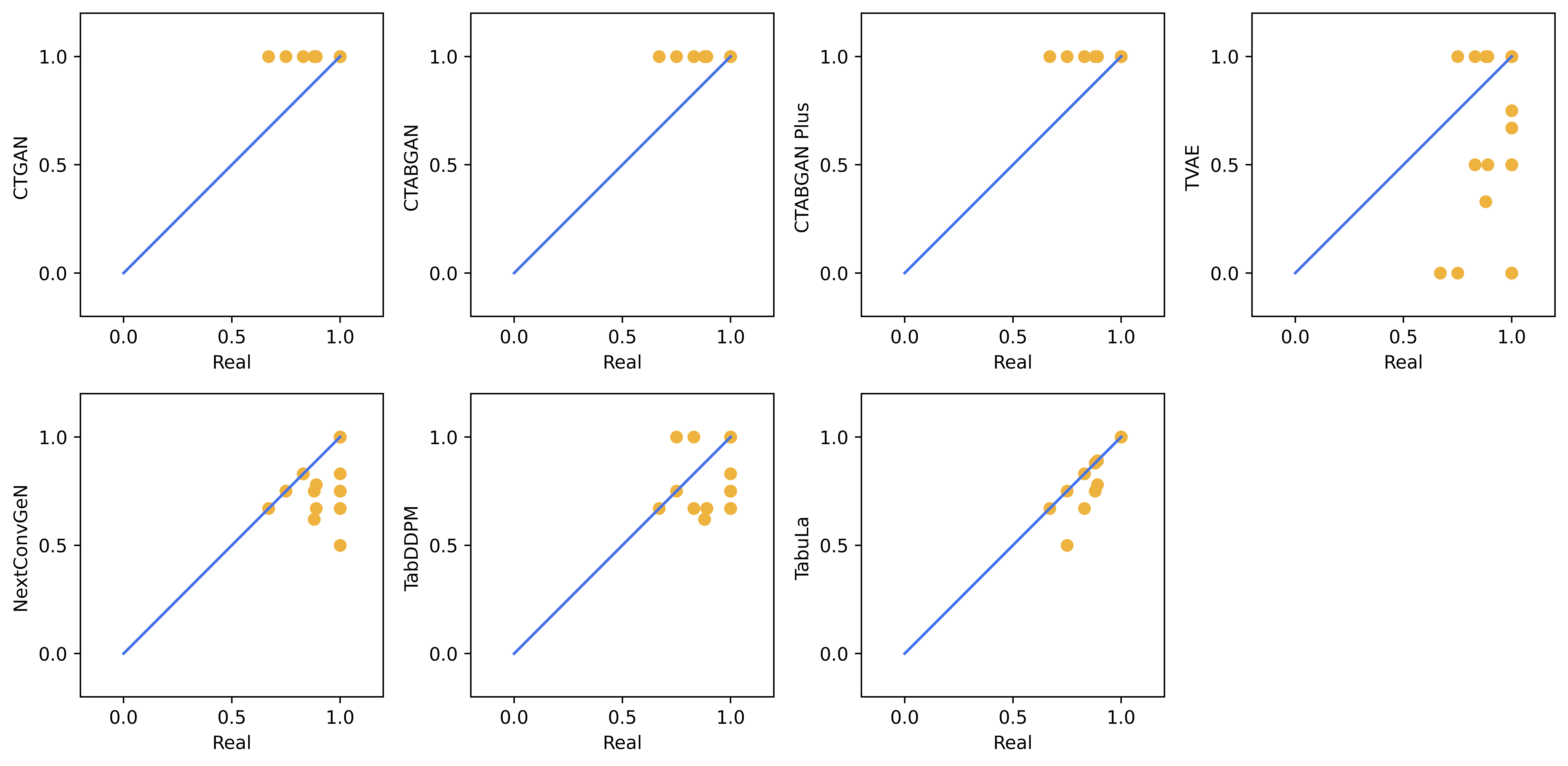}
    \caption{The plots show how closely the $Q$-scores of synthetic Liver Cirrhosis data match those of real Liver Cirrhosis data. Points on the diagonal line signify preserved dependencies, and the points above suggest some real dependencies that are not in synthetic data, while the points below imply that synthetic data introduces new dependencies. Generally, when more points are closer to or on the line, synthetic data better preserves dependencies than real data. \textit{NextConvGeN}, \textit{TabDDPM}, and \textit{TabuLa} effectively capture the logical dependencies in synthetic data, closely reflecting those present in the real data.}
    
    \label{Liver-cirrhosis_Q-metric}
\end{figure}

On the other hand, \textit{NextConvGeN}, \textit{TabDDPM}, and \textit{TabuLa} use real data to generate synthetic samples. \textit{NextConvGeN} produces samples from the convex hull of a sampled data neighborhood, while \textit{TabDDPM} adds noise to real data (forward diffusion) and then iteratively denoises the real data (reverse diffusion) to generate synthetic data.  Additionally, \textit{TabuLa} utilizes real data for model training by converting each row into a format of sentence in a text. Since these approaches have access to real data while generating synthetic data, we argue that they produce superior results in preserving inter-attribute logical dependencies. This can be observed in Figures \ref{migrain_Q-metric}, \ref{airbnb_Q-metric}, \ref{Obesity_Q-metric}, \ref{Stroke_Q-metric}, \ref{Liver-cirrhosis_Q-metric}, which demonstrate inter-attribute logical dependencies between the real data and its respective synthetic data generated using seven generative models.

Figures \ref{migrain_Q-metric}, \ref{airbnb_Q-metric}, \ref{Obesity_Q-metric}, \ref{Stroke_Q-metric}, \ref{Liver-cirrhosis_Q-metric} show how closely the $Q$-scores of the synthetic data align with those of the real data, indicating the preserved dependencies along the diagonal line. Note that in Figure \ref{LD comparison}, for small clinical tabular data (Liver cirrhosis and Migraine data), \textit{TabuLa}, \textit{TabDDPM}, and \textit{NextConvGeN} outperform GAN-based models in preserving inter-attribute logical relationships. A comparison of the $Q$-scores of real and synthetic Liver Cirrhosis data ($418$ samples) from seven generative models is shown in Figure \ref{Liver-cirrhosis_Q-metric}, indicating common dependencies along the diagonal line. \textit{TabuLa}, \textit{TabDDPM}, and \textit{NextConvGeN} demonstrate the unique ability to capture specific logical dependencies, whereas GAN-based and \textit{TVAE} models do not exhibit this capability. Notably, the $Q$-scores of synthetic data are consistently one for all GAN-based models, suggesting that these models failed to capture a single dependency present in real data, possibly due to the limited training data.\par

It is important to note that even if any model preserves $100$\% of the inter-attribute logical dependencies, the $Q$-scores of real and synthetic data may not always align on the diagonal line when plotted. For instance, in the case of Stroke data generated by the \textit{TabDDPM} model, the total number of inter-attribute logical dependencies in real data is $12$. The synthetic Stroke data generated by \textit{TabDDPM} also has $12$ inter-attribute logical dependencies, and all inter-attribute logical dependencies from the real data are preserved in the synthetic data as shown in Figure \ref{LD comparison} (preserved logical dependencies of Stroke data generated by \textit{TabDDPM} is $100$\%). However, when we plot $Q$-scores of real and synthetic data (See \textit{TabDDPM} plot in Figure \ref{Liver-cirrhosis_Q-metric}), only some points fall on the diagonal line, indicating common dependencies. 

\begin{table*}[!ht]
\footnotesize
\begin{center}
\begin{minipage}{0.45\textwidth}
 \begin{tabular}{c|c|c|c} 
  \hline
  
    {\diagbox{B}{A}} & {class $0$} &  {class $1$} &  {class $2$} \\
   \hline
   
    {class $0$} & $0.95$ & $0.04$ & $0$ \\
   \hline
  
    {class $1$} & $0.93$ & $0.05$ & $0.01$ \\
   \hline

    {class $2$} & $0.89$ & $0.08$ & $0.01$ \\
   \hline

    {class $3$} & $0.72$ & $0.16$ & $0.10$ \\
   \hline 
\end{tabular}
\caption{Probabilities of feature A and B in real data}
\label{probability_real}
\end{minipage}
\hspace{0.02\textwidth} 
\begin{minipage}{0.45\textwidth}
\begin{tabular}{c|c|c|c} 
\hline
  
    {\diagbox{B}{A}} & {class $0$} &  {class $1$} &  {class $2$} \\
   \hline
   
    {class $0$} & $1$ & $0$ & $0$ \\
   \hline
  
    {class $1$} & $0.90$ & $0.05$ & $0.01$ \\
   \hline

    {class $2$} & $0.80$ & $0.08$ & $0.03$ \\
   \hline

    {class $3$} & $0.59$ & $0.33$ & $0.07$ \\
   \hline 
\end{tabular}
\caption{Probabilities of feature A and B in synthetic data}
\label{probability_synthetic}
\end{minipage}
\end{center}
\end{table*}

Ideally, we expect all the points to be on the diagonal. This discrepancy occurs because the synthetic data introduces additional dependencies on a particular feature pair absent in the real data, thereby altering the $Q$-scores. The calculation of $Q$-scores depends entirely on the conditional probabilities between feature pairs. For instance, consider features $A$ and $B$, where feature $A$ has three classes and feature $B$ has four. The conditional probabilities of feature pairs in real and synthetic data are provided by Tables \ref{probability_real} and \ref{probability_synthetic}. When we calculate the $Q$-scores for this feature pair in real and synthetic data using Equation \ref{q-function}, we get $0.87$ for real data and $0.75$ for synthetic data. The synthetic data maintains the underlying relationships found in real data, such as the probability of feature $A$ belonging to class $2$ given that feature $B$ belongs to class $0$. However, the $Q$-scores differ due to the existence of additional logical dependencies between attributes in the synthetic data, such as the probability of feature $A$ being in class $0$ or $1$ given that feature $B$ is in class $0$. This is an empirical explanation for why not all data points align perfectly along the diagonal line.\par

Furthermore, mode collapse is a common issue observed in \textit{TVAE} and \textit{CTABGAN Plus} models. This occurs when generative models converge on a limited set of outputs, resulting in the same value for an entire feature, particularly in cases of high imbalance. When a feature has the same value for all rows, it indicates that this feature is functionally dependent on other features in the dataset. In Figure \ref{migrain_Q-metric}, Migraine data generated by \textit{TVAE} exhibited $Q$-scores of $0$ for many feature pairs ($154$ out of $272$), indicating that these feature pairs are functionally dependent. In contrast, these pairs logically depend on real data as the scores lie between $0$ and $1$. Furthermore, \textit{TVAE}'s encoding of information compression in the latent space may limit the decoder's ability to reconstruct all target classes accurately. On the other hand, \textit{NextConvGeN}'s approach involves training generators within each minority sample neighborhood and generating samples accordingly.\par

Our study demonstrates that \textit{TabuLa}, a transformer-based model, outperformed other models in maintaining inter-attribute logical dependencies across various datasets. This is evident from the Figure \ref{LD comparison}. The self-attention mechanism of \textit{TabuLa} enables it to capture more inter-attribute logical dependencies than other models. Generally speaking, self-attention calculates the response at a position in a sequence by attending to all positions within the same sequence. The attention mechanism gives more power to the \textit{TabuLa} to effectively model the inter-attribute logical dependencies in tabular data. Based on our evaluation study, it has been found that \textit{TabuLa}, \textit{TabDDPM}, and \textit{NextConvGeN} are more appropriate choices for handling small and imbalanced datasets in terms of maintaining inter-attribute logical dependencies, as compared to other models.\par

Recent research has primarily focused on modeling tabular data and evaluating its quality using various metrics. However, this study marks the first effort to determine whether generative models can effectively maintain inter-attribute logical and functional dependencies. It is important to note that the research in question only focused on maintaining dependencies for the categorical features within the data, omitting consideration of the continuous features. The study's results highlight the necessity of developing improved model structures that can precisely capture functional and logical relationships. \par

\section{Conclusion}
In this study, for the first time, we evaluate how well synthetic data generated by different generative approaches can preserve the inter-attribute functional and logical dependencies as compared to real data. We introduced a Bayesian logic-based $Q$-function to identify inter-attribute logical dependencies for any given set of features. Using FDTool and the $Q$-function, we examined both functional and logical dependencies in real data and synthetic data generated by seven different generative models for synthetic tabular data generation. The literature shows that none of the synthetic data generation models evaluated the quality of synthetic data in the context of logical and functional dependency preservation. We compared the functional and logical dependencies in the synthetic data with those in real data using five publicly available datasets. The results indicate that \textit{NextConvGeN} and \textit{TabDDPM} models can preserve the logical dependencies in synthetic data for a high proportion of datasets compared to real data, and \textit{TabuLa} can preserve inter-attribute logical dependencies for all the datasets, but none can preserve the functional dependencies from real to synthetic data. This leads us to conclude that if the goal of synthetic data generation is to preserve logical or functional dependencies among features, then there is a necessity for more specialized synthetic data generation models for this purpose. In the field of clinical data analysis, there are various inter-attribute dependencies to consider. Creating synthetic data that accurately reflects real data without revealing sensitive information is important. However, if these dependencies are not preserved in the synthetic data, then such synthetic data might be missing important information or might contain inconsistent information that will affect downstream analyses. We conclude that further research into generative models is necessary to ensure these dependencies are considered when generating samples.

\section*{CRediT authorship contribution statement}
\textbf{Chaithra Umesh:} Experimented and wrote the first version of the manuscript. \textbf{Kristian Schultz:} Worked on the mathematical part and wrote the first version of the manuscript. Both are the joint first authors of this work. \textbf{Manjunath Mahendra:} discussed and revised the manuscript's contents. \textbf{Saptarshi Bej:} Conceptualized, reviewed, and supervised the manuscript writing and experiments. \textbf{Olaf Wolkenhauer:} Reviewed and supervised the manuscript writing and experiments.

\section*{Conflict of Interest}
The authors have no conflict of interest.

\section*{Availability of code and results}
We provided detailed Jupyter notebooks from our experiments in \href{https://github.com/Chaithra-U/Dependency_preservation}{GitHub} to support transparency, re-usability, and reproducibility.

\section*{Acknowledgment}
This work has been supported by the German Research Foundation (DFG), FK 515800538, obtained for `Learning convex data spaces for generating synthetic clinical tabular data'.

\bibliographystyle{elsarticle-num}
\bibliography{main}

\end{document}